\documentclass[runningheads]{llncs}

\usepackage[mobile]{eccv}

\usepackage{eccvabbrv}

\usepackage{graphicx}
\usepackage{booktabs}

\usepackage[accsupp]{axessibility}  

\usepackage{adjustbox}     
\usepackage{multirow}      
\usepackage{makecell}      
\usepackage{pifont}        
\usepackage[table]{xcolor} 

\usepackage{xcolor}
\usepackage{xspace}
\usepackage{caption}
\usepackage{algorithm}
\usepackage{algpseudocode}
\usepackage{amsmath}

\definecolor{baselinecolor}{gray}{.9}
\definecolor{light}{RGB}{179, 224, 255}
\definecolor{darkgreen}{RGB}{0, 150, 0}
\definecolor{darkred}{RGB}{150, 0, 0}

\newcommand{\ours}[1]{\cellcolor{baselinecolor}{#1}}

\usepackage{hyperref}

\usepackage{orcidlink}

\begin{document}

\title{Training-free Uncertainty Guidance for Complex Visual Tasks with MLLMs} 

\titlerunning{Training-free Uncertainty Guidance for Complex Visual Tasks with MLLMs}


\authorrunning{S.~Kim et al.}


\author{Sanghwan Kim\textsuperscript{1,2,3} \qquad Rui Xiao\textsuperscript{1,2} \qquad Stephan Alaniz\textsuperscript{5} \\ Yongqin Xian\textsuperscript{4} \qquad Zeynep Akata\textsuperscript{1,2,3}}

\institute{\textsuperscript{1}Technical University of Munich  \quad \textsuperscript{2}Munich Center for Machine Learning \\
\textsuperscript{3}Helmholtz Munich \,\, \textsuperscript{4}Google \,\, \textsuperscript{5}LTCI, Télécom Paris, Institut Polytechnique de Paris}

\maketitle

\begin{abstract}
Multimodal Large Language Models (MLLMs) often struggle with fine-grained perception, such as identifying small objects in high-resolution images or detecting key moments in long videos. Existing methods typically rely on complex, task-specific fine-tuning, which reduces generalizability and increases system complexity. In this work, we propose an effective, training-free framework that uses an MLLM's intrinsic uncertainty as proactive guidance. Our core insight is that a model's uncertainty decreases when provided with relevant visual information. We introduce a unified mechanism that scores candidate visual inputs by response uncertainty, enabling the model to autonomously focus on the most informative data. We apply this simple principle to three challenging visual tasks: Visual Search, Long Video Understanding, and Temporal Grounding, allowing off-the-shelf MLLMs to achieve performance competitive with specialized, fine-tuned systems. Our results demonstrate that leveraging intrinsic uncertainty is a powerful strategy for improving fine-grained multimodal performance. Code is available
at \url{https://github.com/ExplainableML/ug-framework}.
\end{abstract}    
\section{Introduction}

Multimodal Large Language Models (MLLMs) have achieved notable progress in general visual understanding, yet they often underperform on tasks requiring fine-grained or localized perception~\cite{wu2024v,zou2024seconds,wu2025survey}. A central challenge lies in identifying sparse but crucial information within large and noisy visual contexts. This issue is particularly evident in tasks such as \emph{Visual Search}, which requires detecting small objects in high-resolution images; \emph{Long Video Understanding}, which depends on locating key moments across long video sequences; and \emph{Temporal Grounding}, which involves determining the precise temporal span of an event. In all cases, the vast visual input makes uniform processing impractical, causing models to overlook important details and produce incorrect outputs.

To address this, existing approaches often adopt complex and task-specific solutions. Many involve significant engineering effort, including curating custom datasets for each task and fine-tuning models, or integrating external, specialized models to select relevant regions~\cite{zhang2025thyme, zheng2025deepeyes, li2025dyfo, yu2024frame, huang2024vtimellm}. Although sometimes effective, their high engineering costs and specificity limit generalizability, making them difficult to adapt across diverse tasks requiring nuanced visual reasoning. The core challenge remains: finding a simple, unified, and training-free strategy to guide MLLMs' focus toward the most relevant visual information.

In this work, we propose a principled alternative based on an MLLM's intrinsic uncertainty. While prior work shows that output uncertainty, often measured by entropy, correlates with hallucinations and is useful for post-hoc error detection~\cite{xiao2021hallucination, kadavath2022language, wang2024valid}, its use as a proactive guidance signal has been largely overlooked. We argue that intrinsic uncertainty provides a direct, real-time indicator of which visual input is most informative. Our key insight is that when an MLLM is shown the correct visual evidence for a query, its predictive confidence increases and uncertainty naturally decreases.

Building on this observation, we introduce the \emph{Uncertainty-Guided (UG) framework}, which reformulates diverse localization tasks as a unified problem: finding the state of minimum uncertainty. The framework evaluates candidate visual inputs (e.g., image crops or video frames) and scores them using the MLLM’s response uncertainty. This can be measured via the output distribution's \emph{entropy} or, for binary decisions, a direct \emph{confidence score} from ``yes/no'' probabilities. This simple mechanism enables the model to autonomously identify and attend to the most informative content. Whether selecting the best crop for visual search, sampling the top-$k$ informative frames for video QA, or locating the highest-confidence segment for temporal grounding, the underlying principle remains consistent.

This unified, training-free framework enables off-the-shelf MLLMs to solve complex localization tasks without architectural changes or fine-tuning. We apply it to multiple open-source models, including the LLaVA~\cite{llavaov}, Qwen-VL~\cite{bai2025qwen2}, and InternVL~\cite{chen2024expanding} families. Our contributions are as follows:
(1) We show that minimizing an MLLM’s output uncertainty reliably identifies task‑relevant visual evidence, establishing uncertainty as a principled localization signal.
(2) We introduce a simple, unified, training-free framework that leverages intrinsic uncertainty for diverse fine-grained visual localization tasks without task-specific designs.
(3) Extensive experiments demonstrate that our framework enables standard MLLMs to achieve competitive results with state-of-the-art (SOTA) specialized methods on Visual Search, Long Video Understanding, and Temporal Grounding.
\section{Related Work}

\textbf{Multimodal Large Language Models (MLLMs).}
MLLMs have rapidly advanced into powerful systems combining perception and reasoning. Foundational open-source models such as LLaVA~\cite{liu2023improvedllava,llavaov}, Qwen-VL~\cite{bai2023qwen, bai2025qwen2}, and InternVL~\cite{chen2024internvl, chen2024expanding} effectively align visual features with LLMs. This paradigm has expanded from static images to video with models such as LLaVA-Video~\cite{zhang2024video} and InternVideo~\cite{wang2022internvideo}. Despite progress in general visual question answering, these models still struggle with tasks requiring precise, localized visual reasoning~\cite{wu2024v,zou2024seconds,wu2025survey}.

\noindent
\textbf{Visual Limitations in MLLMs.}
MLLMs exhibit inherent limitations in fine-grained perception. They often fail on small-object recognition due to low-resolution visual encoders that lose critical details~\cite{wu2024v,zhang2025mllms}. Long video understanding remains challenging because restricted LLM context lengths prevent processing full sequences, forcing reliance on suboptimal sampling~\cite{zou2024seconds}. Additionally, most MLLMs lack strong temporal grounding capabilities, as they are not trained with explicit temporal signals or time-aware datasets, thus requiring specialized fine-tuning~\cite{huang2024vtimellm, chen2024timemarker}. Our work addresses all three limitations within a single framework.

\noindent
\textbf{Uncertainty in MLLMs.}
Uncertainty quantification is fundamental for building reliable models. In LLMs and MLLMs, entropy~\cite{shannon1948mathematical} is a key indicator of uncertainty, with higher entropy correlating with hallucinations and incorrect predictions~\cite{kadavath2022language,chen2024context,leng2024mitigating}. Prior work has primarily used uncertainty for post-hoc hallucination detection, whereas we introduce a novel, training-free framework that employs entropy as a proactive signal to guide decision-making in complex visual tasks.
\section{The Uncertainty-Guided (UG) Framework}

In this section, we present the Uncertainty-Guided (UG) framework, a training-free approach that enhances MLLM performance on fine-grained perception tasks. We begin by establishing the core principle underlying our method (\cref{subsec:method_principle}): actively minimizing a model's intrinsic uncertainty serves as an effective strategy for localizing salient visual input. After empirically validating this principle, we formalize the uncertainty metrics used to score visual inputs (\cref{subsec:method_scoring}) and detail how our simple ``score-then-answer'' mechanism applies to three distinct localization challenges (\cref{subsec:method_application}).

\subsection{Principle: Minimizing Uncertainty to Localize Key Visual Information}\label{subsec:method_principle}

Our framework rests on the hypothesis that an MLLM's intrinsic uncertainty can be minimized to localize the key visual information required to answer a query. Whereas prior work uses output entropy for post-hoc error detection, we use it as a proactive localization signal. To evaluate this idea, we conducted a motivating experiment on $V^*$ Bench~\cite{wu2024v}, which contains high-resolution images. Each image was partitioned into square crops with side length equal to one-sixth of $\min(\text{width}, \text{height})$. We fed each crop to LLaVA-OneVision-7B~\cite{llavaov} with the original question, hypothesizing that crops containing salient information would yield \emph{low‑entropy (confident)} predictions, whereas irrelevant crops would yield \emph{high‑entropy (uncertain)} ones.

\begin{figure*}[t]
\vspace{-1em}
\centering
\includegraphics[width=\linewidth]{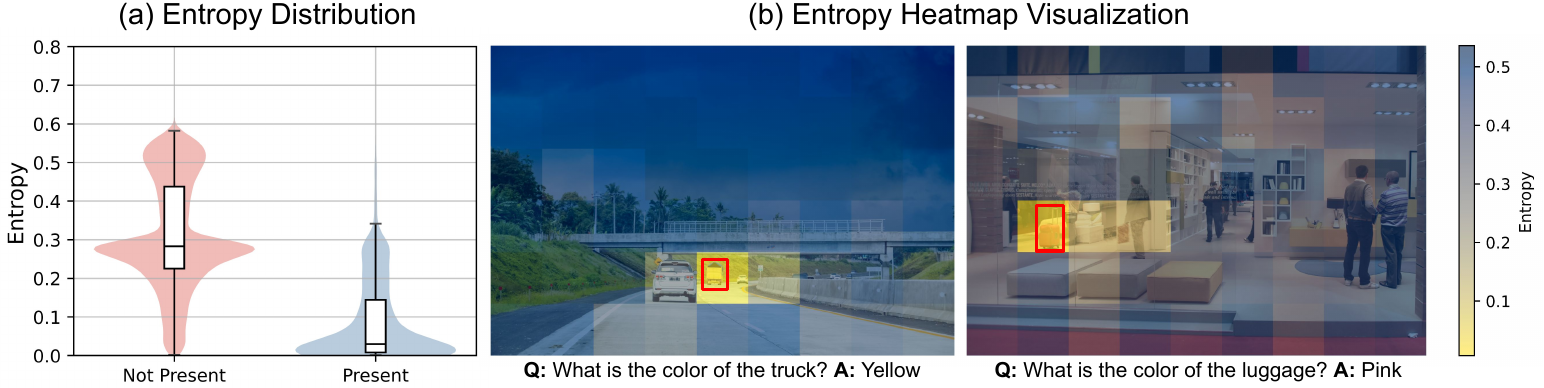}
\caption{(a) \textbf{Entropy distribution} comparing visual crops that include a target object (\texttt{Present}) versus those that do not (\texttt{Not Present}).
(b) \textbf{Entropy heatmap} showing entropy variation across crops. Red box indicates the target object.}
\label{fig:intro}
\end{figure*}

The results in \cref{fig:intro} strongly validate this hypothesis. \cref{fig:intro}a shows a clear, statistically significant separation: \texttt{Present} crops containing the target object (e.g., the truck) exhibit substantially lower entropy than \texttt{Not Present} crops (e.g., background). \cref{fig:intro}b further visualizes this trend, where regions overlapping the target (red box) produce the lowest entropy. This strong correlation confirms that MLLM output uncertainty serves as a reliable proxy for visual relevance. Motivated by this observation, we reformulate fine-grained perception tasks as a search for the visual input that \emph{minimizes model uncertainty}. Additional analysis and theoretical discussion can be found in \cref{supp_sec:correlation,supp_sec:theory}.

\subsection{Uncertainty as a Scoring Function}\label{subsec:method_scoring}

A central component of our framework is quantifying the model's uncertainty for a visual input $v$ and textual query $q$. We use two metrics derived from the output probability distribution $p_i$ at each generation step $i$. For general-purpose relevance scoring, we employ \emph{Average Entropy}. We compute Shannon entropy~\cite{shannon1948mathematical} for each generated token and average it across the sequence of length $T$:
\begin{equation}
\mathcal{H}(v, q) = -\frac{1}{T} \sum_{i=1}^{T} \sum_{j=1}^{N} p_{i,j} \log(p_{i,j})
\label{eq:entropy}
\end{equation}
where $N$ is the vocabulary size and $p_{i,j}$ is the probability of the $j$th vocabulary at step $i$. Lower entropy indicates lower uncertainty and thus higher alignment of the visual input with the query.

For tasks that require a binary decision (e.g., detecting whether an event is present), average entropy may be less informative since confident ``yes'' or ``no'' responses both yield low entropy. In such cases, we use a more targeted metric named \emph{Binary Response Confidence (BRC)} score. It directly measures the model's certainty toward a positive confirmation by taking the difference between the probabilities of the ``yes'' and ``no'' tokens in the first generated distribution $p_1$:
\begin{equation}
    BRC(v, q) = p_1(\text{``yes''}) - p_1(\text{``no''})
    \label{eq:uncertainty}
\end{equation}
This score provides a directional measure of uncertainty, where a high positive value indicates strong confidence in the event's presence.

\begin{figure*}[t]
\centering
\includegraphics[width=\linewidth]{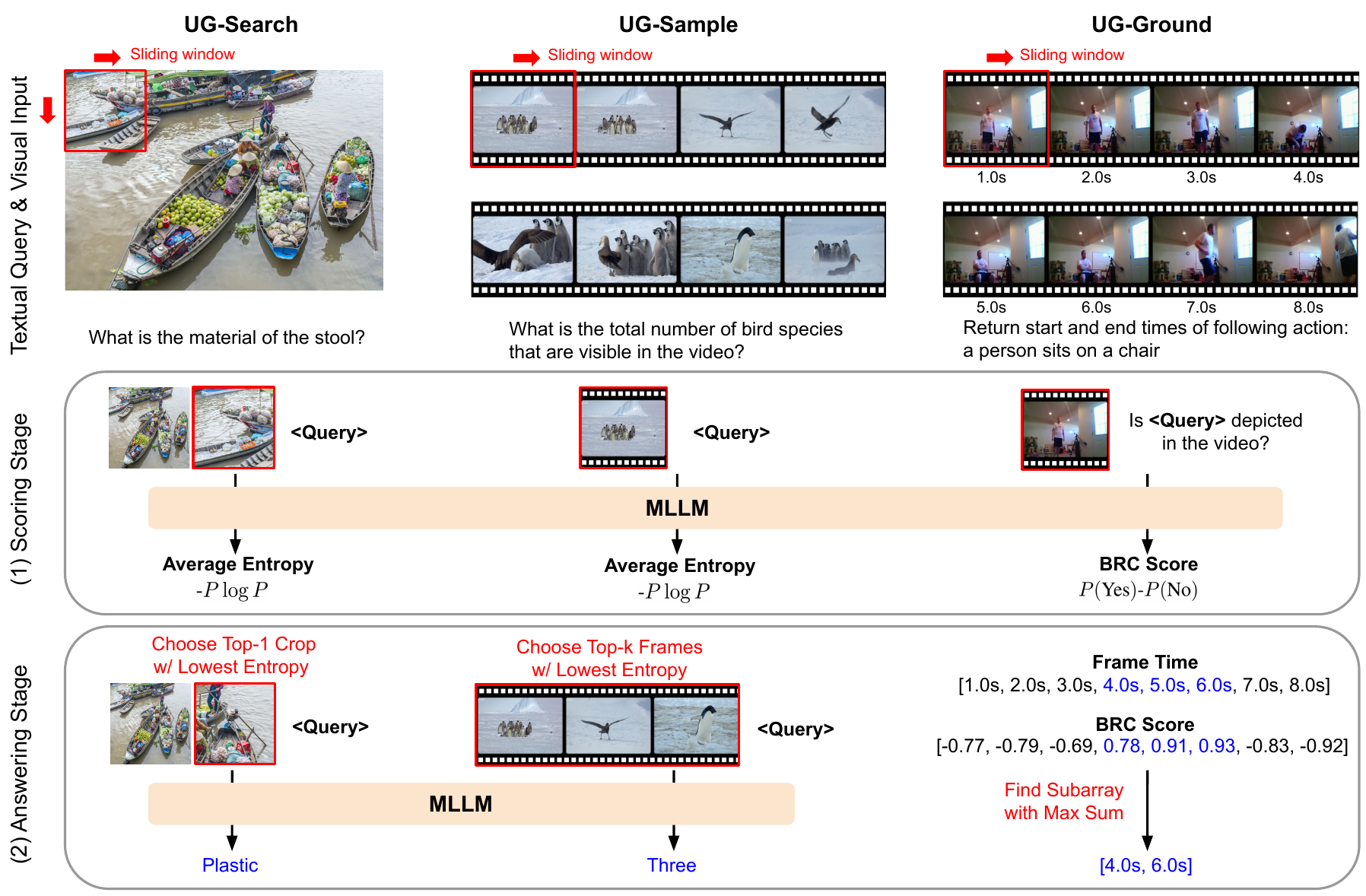}
\caption{\textbf{Uncertainty-Guided (UG) Framework} operates in two stages: (1) Scoring stage: candidate visual inputs (image crops or video frames) are scored using the MLLM's intrinsic uncertainty, measured by either Average Entropy or Binary Response Confidence (BRC) score. (2) Answering stage: the input with the lowest uncertainty is used for a final inference to generate the definitive answer.}
\label{fig:main_figure}
\end{figure*}

\subsection{Applications of the UG Framework}\label{subsec:method_application}
The UG framework's simple ``score-then-answer'' mechanism generalizes across localization tasks including Visual Search, Long Video Understanding, and Temporal Grounding, as shown in \cref{fig:main_figure}.

\noindent
\textbf{UG-Search: Visual Search in High-Resolution Images.}
To localize small objects in large images, UG-Search divides the image into a set of candidate crops using a sliding window. Each crop is scored by feeding it, along with the original image, to the MLLM and calculating its \emph{Average Entropy} (\cref{eq:entropy}). This initial passes are solely for scoring. The single crop that yields the lowest entropy is selected as the most informative region, and the final answer is then generated based on this selected crop only. To improve efficiency in practice, we cache the original image’s key-value (KV) states and reuse them across all scoring passes.

\noindent
\textbf{UG-Sample: Frame Sampling for Long Videos.}
We extend the same principle to long video sampling. To find the most relevant moments, UG-Sample treats each frame (or short window) as a candidate visual input. Each candidate is scored using its \emph{Average Entropy}. Then, the top-$k$ frames with the lowest entropy are selected, concatenated into a single context in temporal order, and used for final inference to answer the query.

\noindent
\textbf{UG-Ground: Temporal Grounding of Events.}
For temporal grounding, UG-Ground reframes the task as finding the most confident contiguous event segment. A sliding window is applied across the video, with each window scored using the \emph{BRC score} (\cref{eq:uncertainty}) by querying whether the event occurs. This yields a sequence of confidence scores. Temporal grounding is thus reduced to finding the subarray with the maximum sum which can be solved efficiently in linear time using \textit{Kadane’s Algorithm}~\cite{bentley1984programming}. The resulting subarray's start and end indices correspond directly to the predicted timestamps.

\section{Experiments}\label{sec:experiment}

We conduct comprehensive experiments to evaluate the effectiveness, generality, and scalability of our Uncertainty-Guided (UG) framework. We assess performance on three challenging localization tasks: Visual Search (\cref{subsec:exp_ug_search}), Video Frame Sampling (\cref{subsec:exp_ug_sample}), and Video Temporal Grounding (\cref{subsec:exp_ug_ground}). Additional analyses and efficiency discussions are presented in \cref{subsec:analysis} and \cref{subsec:efficiency}. All evaluations are performed using \textit{LMMs-Eval}~\cite{zhang2024lmmseval} library for consistency and reproducibility. Implementation details are provided in \cref{supp_sec:details}.

\subsection{UG-Search on Visual Search}~\label{subsec:exp_ug_search}

\begin{table*}[h]
\centering
\vspace{-3em}
\caption{\textbf{Results of UG-Search on Visual Search and Standard QA.} The value in subscript ($\Delta$) indicates the performance gain from the baseline. We also report the average width ($\Bar{W}$) and height ($\Bar{H}$) of the images in each benchmark.}
\label{tab:ug_search}
\setlength\tabcolsep{3pt}
\begin{adjustbox}{width=\textwidth}
\begin{tabular}{l|ccc|cccc}
\toprule
\multirow{2}{*}{\textbf{Model}}  &  \multicolumn{3}{c|}{\textbf{Visual Search QA}} & \multicolumn{4}{c}{\textbf{Standard QA}} \\
& $V^*$ Bench & HR4K & HR8K & TextVQA & POPE & DocVQA & GQA \\
\textit{Image Size ($\Bar{W}\times\Bar{H}$)}  & \textit{2246$\times$1582} & \textit{4023$\times$3502} & \textit{7431$\times$5357}  & \textit{954$\times$818} & \textit{584$\times$478} & \textit{1776$\times$2084} & \textit{578$\times$482} \\
\midrule
LLaVA-OV-7B & 74.4 & 64.9 & 58.4 & 73.8 & 88.4 & 87.1 & 62.3\\
\ours{\quad w/ \textbf{UG-Search}$_{\Delta}$} & \ours{86.9$_{\textcolor{darkgreen}{12.5}}$} & \ours{70.1$_{\textcolor{darkgreen}{5.2}}$} & \ours{68.9$_{\textcolor{darkgreen}{10.5}}$} & \ours{75.1$_{\textcolor{darkgreen}{1.3}}$} & \ours{89.7$_{\textcolor{darkgreen}{1.3}}$} & \ours{88.2$_{\textcolor{darkgreen}{0.9}}$} & \ours{63.0$_{\textcolor{darkgreen}{0.7}}$}\\
Qwen2.5-VL-7B & 64.9 & 60.1 & 53.1 & 76.1 & 86.3 & 93.1 & 60.4 \\
\ours{\quad w/ \textbf{UG-Search}$_{\Delta}$} & \ours{81.7$_{\textcolor{darkgreen}{16.8}}$} & \ours{74.9$_{\textcolor{darkgreen}{14.8}}$} & \ours{66.3$_{\textcolor{darkgreen}{13.2}}$} & \ours{79.7$_{\textcolor{darkgreen}{3.6}}$} & \ours{86.7$_{\textcolor{darkgreen}{0.4}}$} & \ours{94.2$_{\textcolor{darkgreen}{0.9}}$} & \ours{60.7$_{\textcolor{darkgreen}{0.3}}$} \\
InternVL2.5-8B & 71.7 & 59.8 & 58.0 & 77.0 & 90.5 & 91.4 & 62.9 \\
\ours{\quad w/ \textbf{UG-Search}$_{\Delta}$} & \ours{91.1$_{\textcolor{darkgreen}{19.4}}$} & \ours{75.8$_{\textcolor{darkgreen}{16.0}}$} & \ours{73.6$_{\textcolor{darkgreen}{15.6}}$} & \ours{77.7$_{\textcolor{darkgreen}{0.7}}$} & \ours{90.8$_{\textcolor{darkgreen}{0.3}}$} & \ours{92.3$_{\textcolor{darkgreen}{0.9}}$} & \ours{63.4$_{\textcolor{darkgreen}{0.5}}$}\\
\bottomrule
\end{tabular}
\end{adjustbox}
\vspace{-1em}
\end{table*}

\noindent
\textbf{Setup.}
We process each image using a square sliding window. To account for image resolution, we set the crop size to one-sixth of the image for Visual Search QA and to one-half for Standard QA, with the stride set to half the crop size. These hyperparameters are fixed across all benchmarks and are not tuned for individual datasets. UG-Search is applied to several open-source MLLMs, including LLaVA-OneVision~\cite{llavaov}, Qwen2.5-VL~\cite{bai2025qwen2}, and InternVL-2.5~\cite{chen2024expanding}. We evaluate on visual search benchmarks ($V^*$ Bench~\cite{wu2024v} and HR-Bench~\cite{wang2025divide}) and standard benchmarks (TextVQA~\cite{singh2019towards}, POPE~\cite{li2023evaluating}, DocVQA~\cite{mathew2021docvqa}, and GQA~\cite{hudson2019gqa}). Benchmark details appear in \cref{supp_sec:details_of_benchmarks}.

\noindent
\textbf{Results.}
\cref{tab:ug_search} shows that UG-Search consistently improves performance across all base models and datasets. Gains are particularly strong for Visual Search QA, which requires detecting small objects: InternVL2.5 improves by +19.4\% on $V^*$ Bench. This verifies that UG‑Search effectively guides the model toward the correct region of interest. The method also generalizes to Standard QA tasks on regular-size images that still require fine-grained perception. For example, Qwen2.5‑VL gains +3.6\% on TextVQA and +0.9\% on DocVQA, with LLaVA‑OV and InternVL2.5 also showing consistent improvements. These results confirm that UG‑Search enhances fine-grained perception across diverse visual QA settings without any training.

\cref{tab:sota_search} compares UG‑Search with specialized visual search methods built on the same MLLM backbones. Our approach outperforms ZoomEye~\cite{shen2024zoomeye}, another training-free method, across all benchmarks. Methods such as TextCoT~\cite{luan2024textcot} and ViCrop~\cite{zhang2025mllms}, which use the MLLM to generate bounding boxes or rely on attention maps, provide limited gains on high-resolution datasets such as HR-Bench. Although Thyme~\cite{zhang2025thyme} uses supervised fine‑tuning and reinforcement learning to generate cropping code, UG‑Search achieves superior results without training, underscoring the strength of our uncertainty-based approach.

\begin{center}
\vspace{-2em}
\begin{adjustbox}{width=1.0\textwidth}
 {
\begin{minipage}[h]{0.45\linewidth}
\centering
\setlength\tabcolsep{3pt}
\captionof{table}{\textbf{SOTA Comparison of UG-Search.} The \textbf{bold} number represents the best performance.}
    \label{tab:sota_search}
    \resizebox{\linewidth}{!}{%
      \begin{tabular}{lccc}
        \toprule
        Model & $V^*$ Bench & HR4K & HR8K \\
        \midrule
        LLaVA-OV-7B                        & 74.4 & 64.9 & 58.4 \\
        \quad w/ ZoomEye~\cite{shen2024zoomeye} & 85.0 & 68.4 & 66.5 \\
        \ours{\quad w/ \textbf{UG-Search}} & \ours{\textbf{86.9}} & \ours{\textbf{70.1}} & \ours{\textbf{68.9}} \\
        \midrule
        Qwen2.5-VL-7B                      & 64.9 & 60.1 & 53.1 \\
        \quad w/ TextCoT~\cite{luan2024textcot} & 67.0 & 60.6 & 50.6 \\
        \quad w/ ViCrop~\cite{zhang2025mllms}   & 69.6 & 63.3 & 51.9 \\
        \quad w/ Thyme~\cite{zhang2025thyme}    & 68.1 & 66.6 & 59.0 \\
        \ours{\quad w/ \textbf{UG-Search}}      & \ours{\textbf{81.7}} & \ours{\textbf{74.9}} & \ours{\textbf{66.3}} \\
        \bottomrule
      \end{tabular}%
    }
\end{minipage}

\hspace{.03\linewidth}
\begin{minipage}[h]{0.48\linewidth}
\centering
\setlength\tabcolsep{2pt}
\captionof{table}{\textbf{SOTA Comparison of UG-Sample.} The \textbf{bold} number represents the best performance.}
    \label{tab:sota_sample}
\resizebox{\linewidth}{!}{%
      \begin{tabular}{lccc}
        \toprule
        Model & V\,-MME & MLVU & LVB \\
        \midrule
        LLaVA-OV-7B                   & 53.9 & 58.6 & 54.3 \\
        \quad w/ KFC~\cite{fang2025threading}   & 55.4 & \textbf{65.0} & 55.6 \\
        \quad w/ BOLT~\cite{liu2025bolt}        & 56.1 & 63.4 & 55.6 \\
        \quad w/ AKS~\cite{tang2025adaptive}    & 56.1 & 64.8 & 56.8 \\
        \quad w/ FRAG~\cite{huang2025frag}      & 56.3 & 64.9 & 57.3 \\
        \ours{\quad w/ \textbf{UG-Sample}}      & \ours{58.6} & \ours{62.0} & \ours{\textbf{59.5}} \\
        \ours{\quad w/ \textbf{UG-Sample}+AKS}  & \ours{\textbf{59.2}} & \ours{\textbf{65.0}} & \ours{59.4} \\
        \bottomrule
      \end{tabular}%
    }
\end{minipage}
}
\end{adjustbox}
\end{center}

\subsection{UG-Sample on Video Frame Sampling}~\label{subsec:exp_ug_sample}
\vspace{-1em}

\noindent
\textbf{Setup.}
We next evaluate UG framework's ability to identify key query-relevant moments in videos. Following prior work~\cite{liu2025bolt,tang2025adaptive}, we select the top-$8$ frames from a pool of 256 uniformly sampled candidate frames, using a window size of one frame for entropy scoring. For baseline models, 8 frames are uniformly sampled over the entire video. Importantly, these hyperparameters are fixed across all benchmarks, demonstrating that the method does not rely on benchmark‑specific optimization. In addition to previously utilized MLLMs (LLaVA-OneVision, Qwen2.5-VL, and InternVL-2.5), we also apply UG-Sample to video-specialized models (LLaVA-Video~\cite{zhang2024video}, InternVideo-2.5~\cite{wang2025internvideo2}). We benchmark on Long Video QA (Video-MME~\cite{fu2025video}, MLVU~\cite{zhou2025mlvu}, and LongVideoBench~\cite{wu2024longvideobench}) and Short Video QA (EgoSchema~\cite{mangalam2023egoschema}, ActivityNet-QA~\cite{yu2019activitynet}, and NeXT-QA~\cite{xiao2021next}).

\noindent
\textbf{Results.}
\cref{tab:ug_sample} shows that UG‑Sample consistently improves over uniform sampling across all evaluated MLLMs. Notably, the method benefits not only long video tasks but also relatively short video benchmarks, illustrating the broad utility of uncertainty‑guided frame selection. For example, Qwen2.5‑VL gains +6.1\% on Video‑MME and +7.8\% on LongVideoBench. On short‑video datasets, UG‑Sample still provides meaningful improvements, boosting Qwen2.5‑VL by +2.7\% on EgoSchema and +1.6\% on NextQA. As expected, gains are smaller for shorter videos, where uniform sampling is less likely to omit key moments. Overall, these results demonstrate that UG‑Sample is a robust and effective strategy across diverse video lengths.

\begin{table*}[t]
\centering
\caption{\textbf{Results of UG-Sample on Long Video and Short Video QA.} The value in subscript ($\Delta$) indicates the performance gain from the baseline. We also report the average video length of each benchmark.}
\label{tab:ug_sample}
\setlength\tabcolsep{2pt}
\begin{adjustbox}{width=\textwidth}
\begin{tabular}{l|cccc|cc|ccc}
\toprule
\multirow{3}{*}{\textbf{Model}} & \multicolumn{6}{c|}{\textbf{Long Video QA}} & \multicolumn{3}{c}{\textbf{Short Video QA}} \\
 & \multicolumn{4}{c|}{Video-MME \tiny{(w/o sub.)}} & \multirow{2}{*}{MLVU} & \multirow{2}{*}{LVB} & \multirow{2}{*}{EgoSch.} & \multirow{2}{*}{ANQA} & \multirow{2}{*}{NextQA}  \\
 & Overall & Short & Medium & Long & & & & &\\
 \textit{Video Length}  & \textit{17min} & \textit{1.3min} & \textit{9min}  & \textit{41min} & \textit{12min}          & \textit{8min}  & \textit{3min} & \textit{2min} & \textit{0.8min}\\
\midrule
LLaVA-OV-7B & 53.9 & 64.9 & 52.1 & 44.7 & 58.6 & 54.3 & 59.1 & 42.9  & 77.1  \\
\ours{\quad w/ \textbf{UG-Sample}$_{\Delta}$} & \ours{58.6$_{\textcolor{darkgreen}{4.7}}$} & \ours{69.7$_{\textcolor{darkgreen}{4.8}}$} & \ours{58.4$_{\textcolor{darkgreen}{6.3}}$} & \ours{47.7$_{\textcolor{darkgreen}{3.0}}$} & \ours{62.0$_{\textcolor{darkgreen}{3.4}}$} & \ours{59.5$_{\textcolor{darkgreen}{5.2}}$} & \ours{60.3$_{\textcolor{darkgreen}{1.2}}$} & \ours{44.1$_{\textcolor{darkgreen}{1.2}}$} & \ours{77.5$_{\textcolor{darkgreen}{0.4}}$} \\
Qwen2.5-VL-7B    & 53.7 & 62.7  & 51.7 & 46.8 & 53.9 & 52.7 & 53.7 & 44.0  & 69.6  \\
\ours{\quad w/ \textbf{UG-Sample}$_{\Delta}$} & \ours{59.8$_{\textcolor{darkgreen}{6.1}}$} & \ours{69.4$_{\textcolor{darkgreen}{6.7}}$} & \ours{58.9$_{\textcolor{darkgreen}{7.2}}$} & \ours{51.0$_{\textcolor{darkgreen}{4.2}}$} & \ours{55.3$_{\textcolor{darkgreen}{1.4}}$} & \ours{60.5$_{\textcolor{darkgreen}{7.8}}$}& \ours{56.6$_{\textcolor{darkgreen}{2.7}}$} & \ours{45.7$_{\textcolor{darkgreen}{1.7}}$}  & \ours{71.2$_{\textcolor{darkgreen}{1.6}}$}\\
InternVL2.5-8B & 57.8 & 68.1 & 56.4 & 48.8 & 61.6 & 52.8 & 49.2 & 42.6  & 76.6 \\
\ours{\quad w/ \textbf{UG-Sample}$_{\Delta}$} & \ours{60.6$_{\textcolor{darkgreen}{2.8}}$} & \ours{70.3$_{\textcolor{darkgreen}{2.2}}$} & \ours{59.9$_{\textcolor{darkgreen}{3.5}}$} & \ours{51.7$_{\textcolor{darkgreen}{2.9}}$} & \ours{67.6$_{\textcolor{darkgreen}{6.0}}$} & \ours{59.5$_{\textcolor{darkgreen}{6.7}}$} & \ours{50.0$_{\textcolor{darkgreen}{0.8}}$} & \ours{43.6$_{\textcolor{darkgreen}{1.0}}$}  & \ours{77.3$_{\textcolor{darkgreen}{0.7}}$} \\
\midrule
LLaVA-Video-7B & 55.9 & 67.7 & 53.6 & 46.4 & 56.4 & 55.7 & 49.7 & 48.8 & 75.6 \\
\ours{\quad w/ \textbf{UG-Sample}$_{\Delta}$} & \ours{59.7$_{\textcolor{darkgreen}{3.8}}$} & \ours{70.6$_{\textcolor{darkgreen}{2.9}}$} & \ours{60.1$_{\textcolor{darkgreen}{6.5}}$} & \ours{48.3$_{\textcolor{darkgreen}{1.9}}$} & \ours{61.9$_{\textcolor{darkgreen}{5.5}}$} & \ours{58.1$_{\textcolor{darkgreen}{2.4}}$} & \ours{51.2$_{\textcolor{darkgreen}{1.5}}$} & \ours{49.0$_{\textcolor{darkgreen}{0.2}}$} & \ours{77.5$_{\textcolor{darkgreen}{2.9}}$} \\
InternVideo2.5-8B & 54.2 & 62.9 & 52.7 & 47.0 & 59.4 & 51.2 & 56.4 & 48.0 & 76.5 \\
\ours{\quad w/ \textbf{UG-Sample}$_{\Delta}$} & \ours{57.4$_{\textcolor{darkgreen}{3.2}}$} & \ours{65.8$_{\textcolor{darkgreen}{2.9}}$} & \ours{56.0$_{\textcolor{darkgreen}{3.3}}$} & \ours{50.3$_{\textcolor{darkgreen}{3.3}}$} & \ours{63.2$_{\textcolor{darkgreen}{3.8}}$} & \ours{57.7$_{\textcolor{darkgreen}{6.5}}$} & \ours{58.1$_{\textcolor{darkgreen}{1.7}}$} & \ours{49.4$_{\textcolor{darkgreen}{1.4}}$} & \ours{78.2$_{\textcolor{darkgreen}{1.7}}$} \\
\bottomrule
\end{tabular}
\end{adjustbox}
\end{table*}

The comparison in \cref{tab:sota_sample} further demonstrates the strength of our approach. On Video-MME and LongVideoBench, UG-Sample alone outperforms all specialized baselines, including methods that rely on external visual-text similarity scores (BOLT~\cite{liu2025bolt}, AKS~\cite{tang2025adaptive}) and graph-based reasoning (KFC~\cite{fang2025threading}). This suggests that an MLLM's intrinsic uncertainty constitutes a more semantically rich signal for frame relevance than external CLIP similarity or repeated binary prompting (FRAG~\cite{huang2025frag}). On MLVU, however, UG-Sample underperforms relative to competing methods. We attribute this gap to MLVU's \textit{Action Count} category, which demands whole-video coverage: because UG-Sample globally prioritizes the most salient frames, it may under-sample less-prominent target actions distributed across the video. To address this, we construct UG-Sample+AKS, a hybrid that replaces AKS's CLIP-based scoring with our entropy-based score while retaining AKS's segment-wise sampling strategy, thereby ensuring broad temporal coverage. This hybrid achieves state-of-the-art performance on Video-MME and MLVU, confirming that intrinsic model uncertainty and temporal coverage are complementary rather than competing objectives.

\subsection{UG-Ground on Temporal Grounding}~\label{subsec:exp_ug_ground}

\begin{table*}[h]
\centering
\caption{\textbf{Results of UG-Ground on Video Temporal Grounding Benchmarks.} The \textbf{bold} number represents the best performance. The value in subscript ($\Delta$) indicates the improvement over the baseline. The \textbf{bold} number represents the best performance.}
\label{tab:ug_ground}
 \begin{adjustbox}{width=\textwidth}
 {
\setlength\tabcolsep{4pt}
\begin{tabular}{l|cccc|cccc}
\toprule
\multirow{2}{*}{\textbf{Model}}  & \multicolumn{4}{c|}{\textbf{Charades-STA} (\textit{0.5min})} & \multicolumn{4}{c}{\textbf{ActivityNet Captions} (\textit{3min})} \\ 
&  R@0.3 & R@0.5 & R@0.7 & mIoU & R@0.3 & R@0.5 & R@0.7 & mIoU \\
\midrule
\multicolumn{9}{l}{\textit{Instruction-tuned Methods}} \\
TimeChat-7B   & 40.6  & 23.8  & 9.7  & 26.2  & 25.0  & 13.2  & 6.1  & 18.5  \\
VTimeLLM-13B  & 55.3  & 34.3  & 14.7  & 34.6  & 44.8  & 29.5  & 14.2  & 31.4  \\
TimeMarker-8B  & 73.5 & 51.9 & 26.9 & 48.4  & - & - & - & -\\
\midrule
\multicolumn{9}{l}{\textit{Training-free Methods}} \\
VTG-GPT    & 59.5 & 43.7 & 25.9 & 39.8 & 47.1 & 28.3 & 12.8 & 30.5 \\
TFVTG  & 67.0 & 50.0 & 24.3 & 44.5 & 49.3 & 27.0 & 13.4 & 34.1 \\
TAG  & 67.8 & 48.6 & 26.7 & 45.7 & 51.9 & 28.9 & 15.1 & 36.6\\
\midrule
LLaVA-OV-7B & 14.4 & 7.2 & 3.5 & 10.4  & 14.8 & 7.2 & 3.0 & 11.9 \\
\ours{\quad w/ \textbf{UG-Ground}$_\Delta$} & \ours{69.0$_{\textcolor{darkgreen}{54.6}}$} & \ours{45.7$_{\textcolor{darkgreen}{38.5}}$} & \ours{25.9$_{\textcolor{darkgreen}{22.4}}$} & \ours{46.8$_{\textcolor{darkgreen}{36.4}}$} & \ours{53.4$_{\textcolor{darkgreen}{38.6}}$} & \ours{29.2$_{\textcolor{darkgreen}{22.0}}$} & \ours{15.5$_{\textcolor{darkgreen}{12.5}}$} & \ours{37.2$_{\textcolor{darkgreen}{25.3}}$} \\
Qwen2.5-VL-7B & 61.1 & 43.7 & 22.9 & 41.1  & 19.0 & 9.5 & 4.1 & 14.3 \\
\ours{\quad w/ \textbf{UG-Ground}$_\Delta$} & \ours{70.1$_{\textcolor{darkgreen}{9.0}}$} & \ours{51.0$_{\textcolor{darkgreen}{7.3}}$} & \ours{30.1$_{\textcolor{darkgreen}{7.2}}$} & \ours{48.4$_{\textcolor{darkgreen}{7.3}}$} & \ours{54.4$_{\textcolor{darkgreen}{35.4}}$} & \ours{\textbf{31.4}$_{\textcolor{darkgreen}{21.9}}$} & \ours{\textbf{16.5}$_{\textcolor{darkgreen}{12.4}}$} & \ours{37.9$_{\textcolor{darkgreen}{23.6}}$} \\
InternVL2.5-8B  & 11.7 & 4.1 & 1.9 & 9.5  & 6.0 & 2.4 & 1.0 & 5.7 \\
\ours{\quad w/ \textbf{UG-Ground}$_\Delta$} & \ours{67.6$_{\textcolor{darkgreen}{55.9}}$} & \ours{49.3$_{\textcolor{darkgreen}{45.2}}$} & \ours{30.4$_{\textcolor{darkgreen}{28.5}}$} & \ours{47.4$_{\textcolor{darkgreen}{37.9}}$} & \ours{52.7$_{\textcolor{darkgreen}{46.7}}$} & \ours{30.0$_{\textcolor{darkgreen}{27.6}}$} & \ours{16.2$_{\textcolor{darkgreen}{15.2}}$} & \ours{36.9$_{\textcolor{darkgreen}{31.2}}$}\\
\midrule
LLaVA-Video-7B  & 16.8 & 8.3 & 3.0 & 11.4  & 20.9 & 9.9 & 3.9 & 15.4 \\
\ours{\quad w/ \textbf{UG-Ground}$_\Delta$} & \ours{67.5$_{\textcolor{darkgreen}{50.7}}$} & \ours{47.2$_{\textcolor{darkgreen}{38.9}}$} & \ours{29.2$_{\textcolor{darkgreen}{26.4}}$} & \ours{47.2$_{\textcolor{darkgreen}{35.8}}$} & \ours{\textbf{54.9}$_{\textcolor{darkgreen}{34.0}}$} & \ours{30.8$_{\textcolor{darkgreen}{20.9}}$} & \ours{\textbf{16.5}$_{\textcolor{darkgreen}{12.6}}$} & \ours{38.3$_{\textcolor{darkgreen}{22.9}}$} \\
InternVideo2.5-8B  & 50.2 & 32.0 & 14.4 & 32.3  & 18.9 & 9.5 & 4.1 & 14.0 \\
\ours{\quad w/ \textbf{UG-Ground}$_\Delta$} & \ours{\textbf{74.7}$_{\textcolor{darkgreen}{24.5}}$} & \ours{\textbf{52.5}$_{\textcolor{darkgreen}{20.5}}$} & \ours{\textbf{32.6}$_{\textcolor{darkgreen}{18.2}}$} & \ours{\textbf{51.0}$_{\textcolor{darkgreen}{18.7}}$} & \ours{\textbf{54.9}$_{\textcolor{darkgreen}{36.0}}$} & \ours{30.8$_{\textcolor{darkgreen}{21.3}}$} & \ours{16.4$_{\textcolor{darkgreen}{12.3}}$} & \ours{\textbf{38.5}$_{\textcolor{darkgreen}{24.5}}$} \\
\bottomrule
\end{tabular}
}
\end{adjustbox}
\end{table*}

\noindent
\textbf{Setup.} 
Finally, we evaluate our framework on video temporal grounding, which requires identifying the exact start and end times of an event within a video. For baseline, the input is capped at 64 frames due to the context‑length constraints of MLLMs. For UG-Sample, the BRC score is computed using a 15‑frame sliding window with stride 1, and this configuration is used for all datasets without benchmark‑specific hyperparameter tuning. UG‑Ground is applied to the same five MLLMs used in UG‑Sample. Experiments are conducted on two standard benchmarks: Charades‑STA~\cite{gao2017tall} (around 0.5-minute videos) and ActivityNet Captions~\cite{krishna2017dense} (around 3-minute videos). Considering their average video durations, we sample videos at 3 FPS for Charades‑STA and 1 FPS for ActivityNet Captions. Following prior works~\cite{ren2023timechat,chen2024timemarker}, we report standard metrics including Recall@k at various IoU thresholds and mean Intersection over Union (mIoU).

\noindent
\textbf{Results.}
\cref{tab:ug_ground} shows that most MLLMs perform poorly at temporal grounding out of the box as they are not trained for such precise localization. However, UG‑Ground dramatically enhances their capabilities, yielding large improvements across all models and datasets. For example, LLaVA‑OV’s mIoU on Charades‑STA rises from 10.4 to 46.8 (+36.4), demonstrating the substantial impact of uncertainty-guided temporal localization.

Remarkably, this training-free method enables general-purpose MLLMs to outperform two distinct categories of temporal-specialized models: (1) instruction-tuned methods, such as VTimeLLM~\cite{huang2024vtimellm}, TimeChat~\cite{ren2023timechat}, and TimeMarker~\cite{chen2024timemarker} that are explicitly fine-tuned on temporal reasoning data or with time-aware tokens; and (2) training-free methods, such as VTG-GPT~\cite{xu2024vtg}, TFVTG~\cite{zheng2024training}, and TAG~\cite{lee2025tag} that rely on an external pretrained VLM to generate image captions or compute frame-query similarity scores. These results demonstrate that leveraging the MLLM's own internal confidence signal via the BRC score constitutes a more principled and effective strategy for temporal localization than either fine-tuning on temporal reasoning data (VTimeLLM, TimeChat) or introducing time-aware tokens (TimeMarker). UG-Ground also consistently outperforms other training-free approaches such as TFVTG and TAG. Finally, applying UG-Ground to a video-specialized backbone like InternVideo2.5 achieves state-of-the-art performance, underscoring the strength and versatility of our unified grounding framework.

\subsection{UG Framework Analysis}\label{subsec:analysis}
In this section, we conduct a series of analyses to better understand the behavioral properties of the UG framework. We first examine how scoring input granularity affects performance and efficiency, then evaluate scalability with respect to model size, followed by qualitative examples.

\begin{figure*}[t]
\centering
\includegraphics[width=\linewidth]{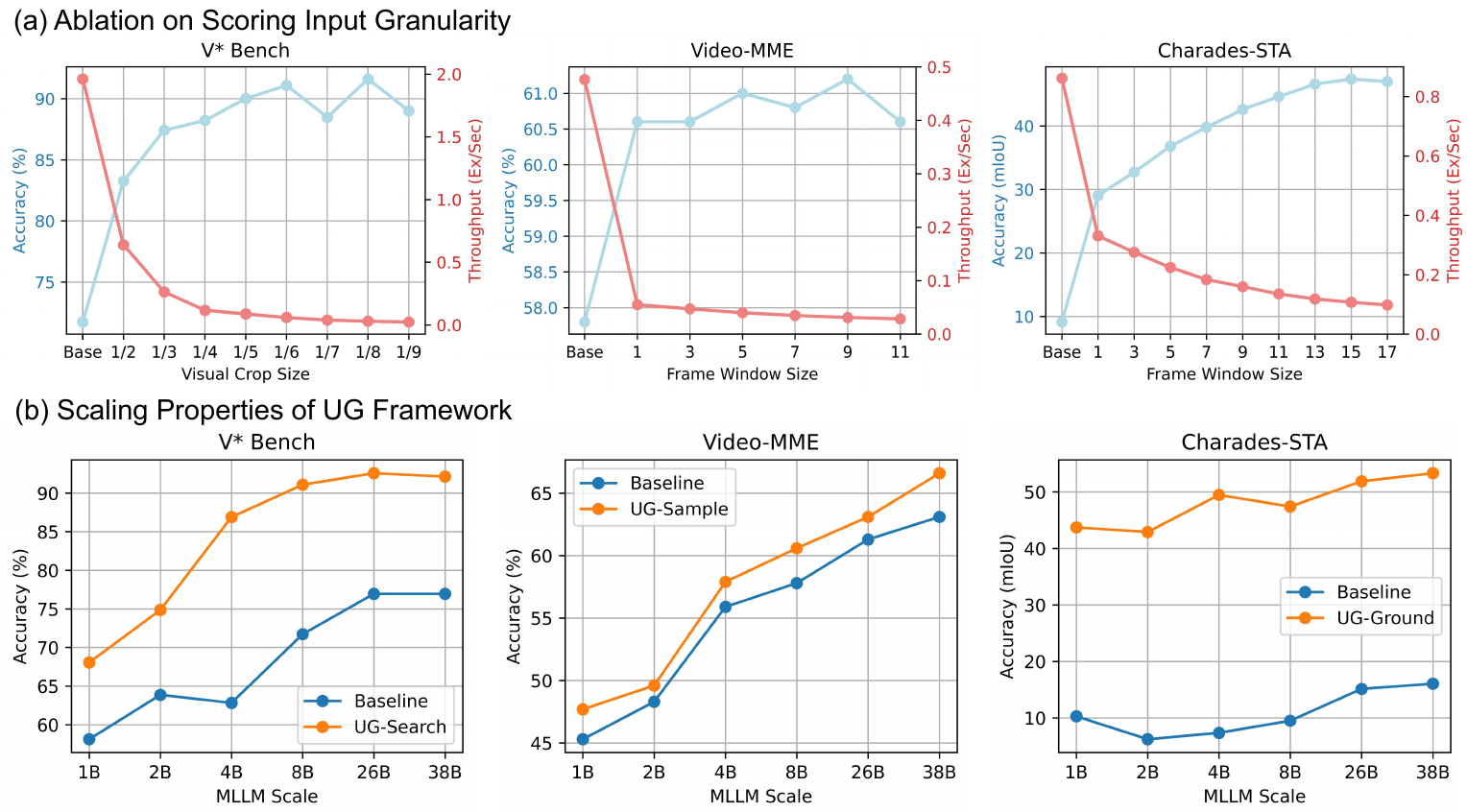}
\vspace{-1em}
\caption{(a) \textbf{Ablation on scoring input granularity} shows a clear trade-off between accuracy and throughput. \texttt{Base} denotes the baseline. (b) \textbf{Scaling properties of UG framework} show consistent performance gains as InternVL‑2.5 model size increases.}\label{fig:ablation}
\vspace{-1em}
\end{figure*}

\begin{figure*}[t!] 
\centering
\includegraphics[width=\linewidth]{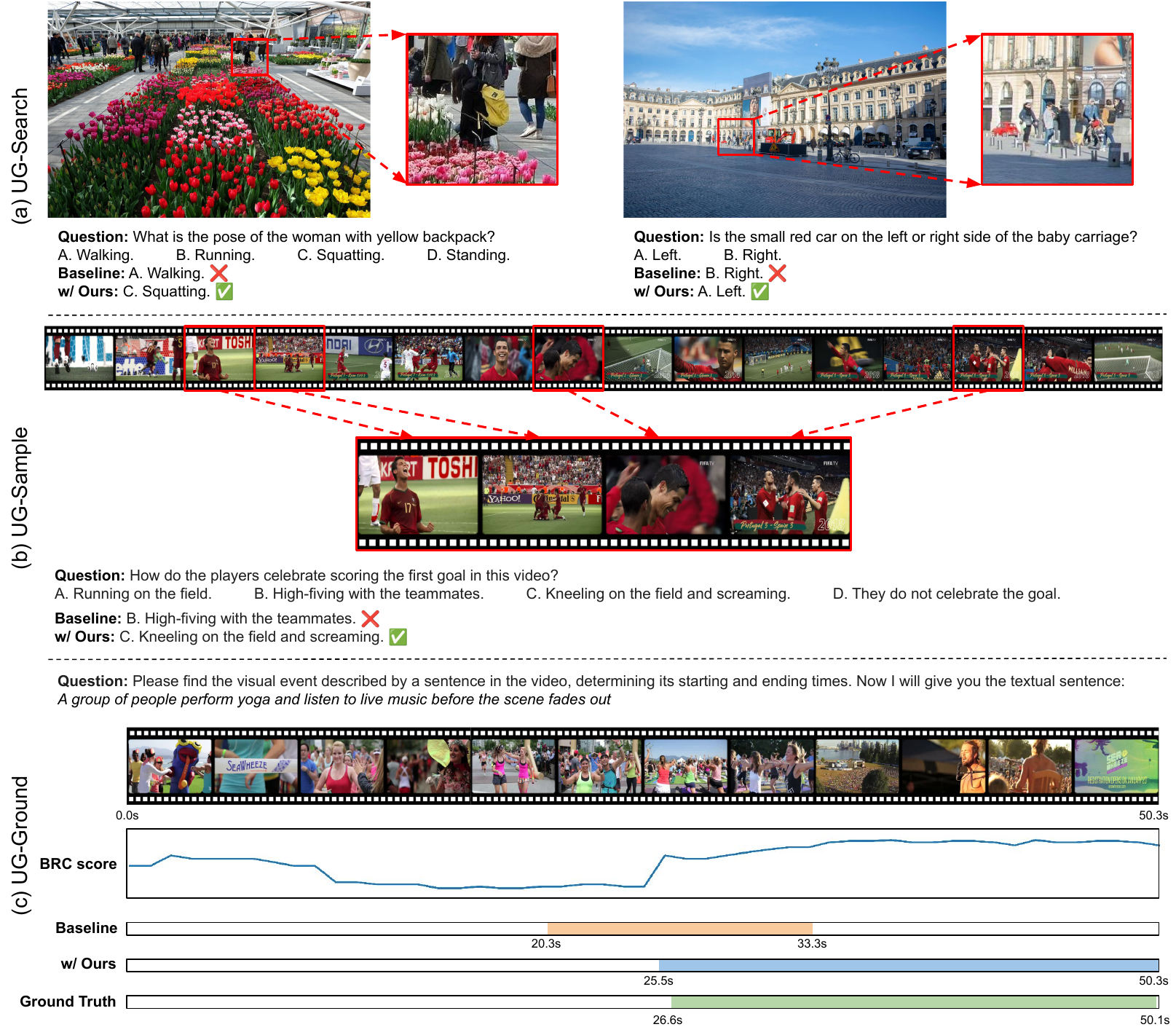}
\vspace{-1em}
\caption{\textbf{Qualitative Results} with InternVL2.5-8B as the baseline. (a) UG-Search localizes small target objects by selecting the lowest-entropy crop (red box). (b) UG-Sample identifies key semantic frames (red box) with the lowest entropy from a long video understanding. (c) UG-Ground pinpoints the correct event timeline by finding the peak in its BRC score sequence.}
\label{fig:qualitative}
\vspace{-1em}
\end{figure*}

\noindent
\textbf{Ablation on Scoring Input Granularity.}
We first examine how the granularity of candidate visual inputs used during scoring (crop size or window size) affects accuracy and efficiency using InternVL2.5‑8B. Throughput (examples/second) is measured on NVIDIA H100 GPUs. As shown in \cref{fig:ablation}a, results exhibit a clear trade‑off between accuracy and speed.

For UG‑Search on $V^*$ Bench, accuracy (light blue) peaks at finer granularity (crop size of $1/8$), with gains saturating beyond $1/6$. Throughput (red), however, declines sharply as granularity increases, dropping from roughly 0.6 at $1/2$ to under 0.1 at $1/9$. This shows that extremely fine crops yield diminishing accuracy gains while incurring substantial computational cost.

For video tasks, the optimal granularity varies by task. UG‑Ground evaluated on Charades‑STA benefits from larger temporal windows, achieving its best performance with a 15‑frame window. In contrast, UG‑Sample on Video‑MME achieves near‑optimal accuracy with a single frame, as capturing isolated key moments matters more than longer temporal context. In both cases, throughput decreases almost linearly with respect to window size.

Across all granularities, UG consistently surpasses the baseline, demonstrating robustness. In practice, users can select coarser inputs (e.g., $1/2$ crops or 1‑frame windows) for higher throughput, or finer inputs (e.g., $1/8$ crops or 15‑frame windows) for maximum accuracy, offering flexible control over the performance–efficiency trade‑off.

\noindent
\textbf{Scaling Properties of UG Framework.}
\cref{fig:ablation}b evaluates the scalability of the UG across the InternVL‑2.5 model family, spanning 1B to 38B parameters. Both the original baseline models and their UG‑enhanced variants improve steadily as model size increases. Crucially, the benefits of UG remain stable and consistent across the entire scaling curve, and larger models achieve the highest absolute performance. This consistent improvement suggests that as MLLMs become larger and, presumably, better calibrated, their intrinsic uncertainty becomes an even more reliable signal that our framework can effectively harness.

\noindent
\textbf{Qualitative Results.}
\cref{fig:qualitative} provides qualitative examples demonstrating how the UG framework isolates the most relevant visual information in noisy visual contexts, enabling correct predictions where baseline models fail. The examples are drawn from $V^*$ Bench, Video-MME, and ActivityNet Captions, with additional qualitative results included in the \cref{supp_sec:qualitative}.

In the examples for (a), UG-Search overcomes the baseline's tendency to be distracted by cluttered scenes. In the left image, it isolates the woman with the backpack to correctly identify her pose as ``Squatting'', while the baseline incorrectly guesses a common pose (``Walking''). Similarly, on the right image, it successfully localizes ``small red car'' and ``baby carriage'' to accurately resolve their spatial relationship. 

The video example in (b) shows UG-Sample identifying the precise frames of a goal celebration. The baseline, likely processing irrelevant moments from a uniform sample, fails to understand the event correctly. Our method, however, filters the timeline to provide the necessary context relevant to the query, leading to the correct answer (``C. Kneeling on the field and screaming''). 

Finally, in (c), the baseline fails to identify the correct time segment for the ``yoga and live music'' event. Our UG-Ground method first transforms the video into a sequence of BRC scores, which exhibits a clear plateau of high confidence during the correct interval. This allows our method to accurately ground the event's duration, aligning almost perfectly with the ground truth.

\subsection{Improving Efficiency of UG Framework}\label{subsec:efficiency}

The main limitation of the UG framework is its increased runtime due to the multiple scoring passes required during evaluation. To address this, we introduce three practical strategies for improving efficiency: (1) decoupling the scoring and answering MLLMs, (2) leveraging external vision models as pre‑filters, and (3) dynamically adjusting the stride for temporal windows. These approaches offer practical avenues for significantly reducing computational cost while maintaining strong performance. Moreover, the UG framework is inherently parallelizable: crops or frames can be processed independently, enabling substantial reductions in wall-clock time when batched or distributed across devices.

\begin{table}[h]
\centering
\caption{\textbf{Results with Decoupled Scoring and Answering Models.} Using a larger MLLM for scoring improves accuracy, while a smaller scorer increases throughput during inference. Throughput is reported in example per second.}
\label{tab:ug_scorer_infer}
 \begin{adjustbox}{width=0.9\textwidth}
 {
\setlength\tabcolsep{6pt}
\begin{tabular}{ll|cc|cc}
\toprule
\multicolumn{1}{c}{\textbf{Answering}} & \multicolumn{1}{c|}{\textbf{Scoring}} & \multicolumn{2}{c|}{\textbf{UG-Search}} & \multicolumn{2}{c}{\textbf{UG-Sample}} \\ 
\multicolumn{1}{c}{\textbf{MLLM}} & \multicolumn{1}{c|}{\textbf{MLLM}} &  $V^*$ Bench &  Throughput  & Video-MME & Throughput \\
\midrule
InternVL2.5-8B & None & 71.7 & 1.961 & 57.8 & 0.476 \\\midrule
InternVL2.5-8B & InternVL2.5-1B & 75.4 & \textbf{0.139} & 58.5 & \textbf{0.107} \\
InternVL2.5-8B & InternVL2.5-2B & 79.1 & 0.097 & 59.0 & 0.097 \\
InternVL2.5-8B & InternVL2.5-4B & 89.0 & 0.078 & 60.1 & 0.070 \\
InternVL2.5-8B & InternVL2.5-8B & 91.1 & 0.060 & 60.6 & 0.054 \\
InternVL2.5-8B & InternVL2.5-26B & \textbf{92.0} & 0.037 & \textbf{61.8} & 0.036 \\
\bottomrule
\end{tabular}
}
\end{adjustbox}
\end{table}

\noindent
\textbf{Decoupling the Scoring and Answering Models.}
We explore separating the MLLM used during the uncertainty-scoring stage from the model used for final answer generation. As shown in \cref{tab:ug_scorer_infer}, all decoupled configurations yield clear gains over the no-scoring baseline (71.7\% and 57.8\% on $V^*$ Bench and Video-MME, respectively), confirming that the scoring step itself is the primary driver of improvement. Across scoring model sizes, accuracy and throughput trade off monotonically: a 1B scorer achieves the highest throughput among UG variants (0.139 on $V^*$ Bench) but the lowest accuracy (75.4\%), while a 26B scorer delivers the best accuracy on both benchmarks ($V^*$ Bench: 92.0\%, Video-MME: 61.8\%) at the cost of reduced throughput (0.037). Notably, the 4B scorer reaches 89.0\% on $V^*$ Bench within 2.1 points of self-scoring with the 8B model while maintaining a $1.3\times$ throughput advantage, representing a favorable accuracy--efficiency operating point. The consistent accuracy gain from larger scorers aligns with the finding in \cref{subsec:analysis} that larger, better-calibrated models produce more reliable entropy signals. This decoupling strategy opens practical directions such as using a lightweight distilled scorer or a cascaded scoring scheme to match large-scorer accuracy at lower cost.

\begin{table*}[h]
  \centering
  \caption{\textbf{Results of Utilizing External Vision Models as Pre-filters.}}
  \vspace{-1em}
  \setlength\tabcolsep{5pt}
  \renewcommand{\arraystretch}{1.1}
  \begin{subtable}[h]{0.48\linewidth}
    \centering
    \scriptsize
    \caption{\textbf{UG-Search with Pre-filters.} UG-Search is applied to smaller subsets of visual candidates proposed by YOLOv8 and SAM2.}
    \label{tab:search_external}
    \resizebox{\linewidth}{!}{%
        \begin{tabular}{lcc}
        \toprule
           Method                    & $V^*$ & Throughput \\
        \midrule
        LLaVA-OV-7B            & 74.4  & 1.951 \\
        \ours{\quad w/ \textbf{UG-Search}} & \ours{\textbf{81.2}} & \ours{0.420} \\
        \ours{\qquad + YOLOv8 Pre-filter} & \ours{74.9} & \ours{\textbf{0.678}} \\
        \ours{\qquad + SAM2 Pre-filter}   & \ours{78.2} & \ours{0.228} \\
        \bottomrule
        \end{tabular}
    }
  \end{subtable}
  \hfill
  \begin{subtable}[h]{0.48\linewidth}
    \centering
    \scriptsize
    \caption{\textbf{UG-Sample with Pre-filters.} Top-32 frames are pre-filtered by SigLIP and DINOv2 before UG-Sample.}
    \label{tab:sample_external}
    \resizebox{\linewidth}{!}{%
        \begin{tabular}{lcc}
        \toprule
         Method                      & V-MME & Throughput \\
        \midrule
        LLaVA-OV-7B            & 53.9  & 1.048 \\
        \ours{\quad w/ \textbf{UG-Sample}} & \ours{\textbf{58.6}} & \ours{0.130} \\
        \ours{\qquad + DINOv2 Pre-filter} & \ours{56.4} & \ours{\textbf{0.232}} \\
        \ours{\qquad + SigLIP Pre-filter} & \ours{57.3} & \ours{0.210} \\
        \bottomrule
        \end{tabular}
    }
  \end{subtable}
\end{table*}

\noindent
\textbf{Using External Pre-filters to Improve Efficiency.} 
We incorporate lightweight, off‑the‑shelf vision models as efficient \textit{pre-filters} to reduce the number of candidates that the UG framework must score. These models identify a smaller and more promising subset of regions or frames, thereby lowering computational cost.

For UG-Search, we use region proposals from YOLOv8~\cite{varghese2024yolov8} and segmentation masks from SAM2~\cite{ravi2024sam} to construct a reduced set of candidate crops. As shown in \cref{tab:search_external}, full UG-Search achieves the highest accuracy (81.2\%), albeit with a significant throughput reduction relative to the vanilla MLLM (0.420 vs.\ 1.951). Applying YOLOv8 as a pre-filter improves throughput among UG-Search variants (0.678) but yields only marginal gains over the vanilla MLLM (74.9\% vs.\ 74.4\%). SAM2 improves accuracy to 78.2\% but incurs the highest latency (0.228), making its efficiency–accuracy trade-off less favorable than full UG-Search. We attribute this to the query-agnostic nature of generic detectors: without access to the language query, they tend to miss small or context-dependent targets, a finding also reported in ViCrop~\cite{zhang2025mllms}. Overall, these results suggest that in complex scenes, exhaustive UG-Search remains more reliable than generic region-selection pre-filters.

In contrast, pre-filtering offers a meaningful throughput--accuracy trade-off for video tasks. As shown in \cref{tab:sample_external}, both DINOv2~\cite{oquab2023dinov2} and SigLIP~\cite{zhai2023sigmoid} pre-filters substantially improve throughput over full UG-Sample (0.232 and 0.210 vs.\ 0.130) while retaining most of the accuracy gain over the baseline. DINOv2 achieves the highest throughput ($1.8\times$ over UG-Sample) but at a larger accuracy cost (56.4\% vs.\ 58.6\%). SigLIP offers a better accuracy--efficiency balance: it narrows the accuracy gap to only 1.3 points below full UG-Sample (57.3\% vs.\ 58.6\%) while still delivering a $1.6\times$ throughput improvement, and remains well above the no-sampling baseline (53.9\%). This indicates that query-aware semantic pre-filtering can recover a significant portion of the compute overhead introduced by UG-Sample with only modest accuracy degradation.

\begin{table*}[h]
  \centering
  \caption{\textbf{Results of Video Frame and Stride Adjustment.}}
  \vspace{-1em}
  \setlength\tabcolsep{2pt}
  \renewcommand{\arraystretch}{1.1}
  \begin{subtable}[h]{0.48\linewidth}
    \centering
    \scriptsize
    \caption{\textbf{UG-Sample} evaluated on Video-MME. We ablate Frame $\in\{1,9\}$ and Stride $\in\{1,3,5,7,9\}$ during scoring stage. }\label{tab:supp_sample_ablation}
    \resizebox{\linewidth}{!}{
    \begin{tabular}{lcc}
    \toprule       
    Method & V-MME & Throughput\\\midrule
    InternVL2.5-8B    &    57.8  &  0.476    \\
    \ours{\quad w/ Frame=1, Stride=1}        &    \ours{60.6}   &  \ours{0.054}   \\
    \ours{\quad w/ Frame=9, Stride=1}      &    \ours{61.1}   &  \ours{0.025}   \\
    \ours{\quad w/ Frame=9, Stride=3}        &    \ours{\textbf{62.2}}   &  \ours{0.051}   \\
    \ours{\quad w/ Frame=9, Stride=5}        &    \ours{61.9}   &  \ours{0.067}   \\
    \ours{\quad w/ Frame=9, Stride=7}        &    \ours{61.3}   &  \ours{0.086}   \\
    \ours{\quad w/ Frame=9, Stride=9}  &    \ours{60.6}   &  \ours{\textbf{0.135}}   \\
    \bottomrule 
    \end{tabular}
    }
  \end{subtable}
  \hfill
  \begin{subtable}[h]{0.48\linewidth}
    \centering
    \scriptsize
    \caption{\textbf{UG-Ground} evaluated on Charades-STA. We ablate Frame $\in\{7,15\}$ and Stride $\in\{1,3,5,7,9,11\}$ during scoring stage.}\label{tab:supp_ground_ablation}
    \resizebox{\linewidth}{!}{
    \begin{tabular}{lcc}
    \toprule     
    Method & Charades & Throughput\\\midrule
    InternVideo2.5-8B    &    32.3  &  1.370    \\
    \ours{\quad w/ Frame=7, Stride=1}       &    \ours{45.5}   &  \ours{0.205}   \\
    \ours{\quad w/ Frame=15, Stride=1}      &    \ours{\textbf{51.0}}   &  \ours{0.150}   \\
    \ours{\quad w/ Frame=15, Stride=3}       &    \ours{\textbf{51.0}}   &  \ours{0.403}   \\
    \ours{\quad w/ Frame=15, Stride=5}       &    \ours{50.9}   &  \ours{0.599}   \\
    \ours{\quad w/ Frame=15, Stride=7}        &    \ours{50.6}   &  \ours{0.787}   \\
    \ours{\quad w/ Frame=15, Stride=9}       &    \ours{50.2}   &  \ours{0.877}   \\
    \ours{\quad w/ Frame=15, Stride=11}      &    \ours{49.7}   &  \ours{\textbf{1.064}}   \\
    \bottomrule 
    \end{tabular}
    }
  \end{subtable}
\end{table*}

\noindent
\textbf{Dynamic Stride Adjustment for Frame-Window.}
We introduce dynamic stride control to balance temporal resolution with computational efficiency in both UG-Sample and UG-Ground. Increasing the stride reduces window overlap and thus lowers the number of forward passes required during the scoring stage of video-based UG methods.

\cref{tab:supp_sample_ablation} reports results on Video-MME using InternVL2.5-8B. The default UG-Sample configuration corresponds to Frame=1 and Stride=1. We then fix the window size to 9 frames and vary the stride, selecting the center frame of each window (e.g., the 5th frame in a 9-frame window) for Top-$K$ frame selection. Increasing the stride from 1 to 9 increases throughput dramatically (0.025 vs.\ 0.135) while still outperforming the baseline (60.6\% vs.\ 57.8\%), showing that UG-Sample is highly robust to temporal sparsification.

A similar pattern holds for UG‑Ground. The stride controls the temporal density of uncertainty scoring, governing the trade‑off between localization precision and computational cost. As shown in \cref{tab:supp_ground_ablation}, the default configuration (Frame=15, Stride=1) delivers the strongest performance at 51.0 mIoU. Increasing the stride introduces only minor degradation: a stride of 3 reaches the same peak accuracy, and even a stride of 11 maintains strong performance at 49.7 mIoU. Throughput, however, improves substantially (0.150 → 1.064), adding only modest overhead relative to the baseline while preserving a large accuracy advantage (49.7 vs.\ 32.3 mIoU).

\section{Conclusion and Limitation}
In this work, we introduced the Uncertainty‑Guided (UG) framework, a training‑free approach that leverages an MLLM’s intrinsic uncertainty to address challenging fine‑grained perception tasks, including Visual Search, Long Video Understanding, and Temporal Grounding. Our experiments show that this simple yet principled strategy enables standard MLLMs to achieve performance competitive with heavily fine‑tuned, task‑specific systems, demonstrating that uncertainty minimization is an effective and efficient mechanism for guiding multimodal reasoning.

\noindent
\textbf{Limitation.} The primary limitation of our framework is the increased inference cost introduced by the scoring stage, which requires multiple forward passes. However, this overhead can be alleviated using the efficiency strategies discussed in Sec.~\ref{subsec:efficiency}. More broadly, the UG framework aligns with the growing paradigm of test‑time scaling, trading additional computation for enhanced capability, but differs in that its computation is inherently parallelizable: image crops or video frames are independent units that can be batched or distributed across devices to reduce wall‑clock latency. Although our current implementation performs scoring sequentially due to resource and stability constraints, a fully parallelized design would substantially accelerate inference. We leave this direction, along with further efficiency optimizations, for future work.

\section*{Acknowledgements.} 
This work was partially funded by the ERC (853489 - DEXIM) and the Alfried Krupp von Bohlen und Halbach Foundation, which we thank for their generous support. We also gratefully acknowledge support from the German Research Foundation (DFG) via SFB 1233 ``Robust Vision: Inference Principles and Neural Mechanisms,'' TP A2 (Project No. 276693517), and from Hi! PARIS and the ANR/France 2030 program (ANR-23-IACL-0005).


%
%
\bibliographystyle{splncs04}
\bibliography{main}

\newpage
\appendix

\begin{center}
  {\large\bfseries Supplementary Materials}
\end{center}
\vspace{0.5em}

\noindent\textbf{Contents}\\[0.4em]
\noindent\S A\quad \hyperref[supp_sec:interpretation]{Interpretation of Uncertainty in MLLMs}\\
\noindent\S B\quad \hyperref[supp_sec:correlation]{The Inverse Correlation Between Performance and Entropy}\\
\noindent\S C\quad \hyperref[supp_sec:theory]{Theoretical Discussion of the UG Framework}\\
\hspace*{2em}C.1\enspace \hyperref[supp_subsec:notation]{Notation and Setup}\\
\hspace*{2em}C.2\enspace \hyperref[supp_subsec:ib_background]{Background: The Information Bottleneck Objective}\\
\hspace*{2em}C.3\enspace \hyperref[supp_subsec:main_theorem]{UG Framework as Inference-Time IB Optimization}\\
\noindent\S D\quad \hyperref[supp_sec:details_of_baselines]{Details of Baseline Methods}\\
\hspace*{2em}D.1\enspace \hyperref[supp_subsec:general_baselines]{General MLLM Baselines}\\
\hspace*{2em}D.2\enspace \hyperref[supp_subsec:baselines_search]{Visual Search}\\
\hspace*{2em}D.3\enspace \hyperref[supp_subsec:baselines_sample]{Video Frame Sampling}\\
\hspace*{2em}D.4\enspace \hyperref[supp_subsec:baselines_ground]{Temporal Grounding}\\
\noindent\S E\quad \hyperref[supp_sec:details_of_benchmarks]{Details of Benchmarks}\\
\hspace*{2em}E.1\enspace \hyperref[supp_subsec:bench_search]{Visual Search}\\
\hspace*{2em}E.2\enspace \hyperref[supp_subsec:bench_sample]{Video Frame Sampling}\\
\hspace*{2em}E.3\enspace \hyperref[supp_subsec:bench_ground]{Temporal Grounding}\\
\noindent\S F\quad \hyperref[supp_sec:details]{Implementation Details}\\
\hspace*{2em}F.1\enspace \hyperref[supp_subsec:ug_search_impl]{UG-Search}\\
\hspace*{2em}F.2\enspace \hyperref[supp_subsec:ug_sample_impl]{UG-Sample}\\
\hspace*{2em}F.3\enspace \hyperref[supp_subsec:ug_ground_impl]{UG-Ground}\\
\noindent\S G\quad \hyperref[supp_sec:extra_ablation]{Extra Ablation Studies}\\
\hspace*{2em}G.1\enspace \hyperref[supp_subsec:topk_ablation]{Ablation on Top-$K$ Crops of UG-Search}\\
\hspace*{2em}G.2\enspace \hyperref[supp_subsec:scoring_ablation]{Ablation on Scoring Methods}\\
\hspace*{2em}G.3\enspace \hyperref[supp_subsec:entropy_precision]{Precision of the Entropy Signal}\\
\hspace*{2em}G.4\enspace \hyperref[supp_subsec:test_time_compute]{Comparison with Test-Time Compute Baselines}\\
\noindent\S H\quad \hyperref[supp_sec:extra_results]{Extra Results of UG Framework}\\
\hspace*{2em}H.1\enspace \hyperref[supp_subsec:extra_ug_search]{Extra Results with UG-Search}\\
\hspace*{2em}H.2\enspace \hyperref[supp_subsec:extra_ug_sample]{Extra Results with UG-Sample}\\
\noindent\S I\quad \hyperref[supp_sec:qualitative]{Extra Qualitative Results}\\

\section{Interpretation of Uncertainty in MLLMs}\label{supp_sec:interpretation}
Quantifying uncertainty is a cornerstone of building reliable and interpretable machine learning systems~\cite{ghahramani2015probabilistic,gal2016dropout,lakshminarayanan2017simple}. Entropy, a foundational concept from information theory~\cite{shannon1948mathematical}, provides a formal measure of the uncertainty inherent in a probability distribution. In the context of machine learning, entropy has long been integral to various algorithms and objective functions. For instance, it underpins the construction of decision trees and serves as the basis for the cross-entropy loss function, a standard for training deep neural networks in classification tasks~\cite{goodfellow2016deep}. These applications leverage entropy to assess prediction confidence, guide model optimization, and provide insights into model behavior.

The advent of large language models (LLMs)~\cite{devlin2019bert,brown2020language,touvron2023llama} has renewed interest in entropy as a vital tool for understanding and refining model performance~\cite{shannon1951prediction,kadavath2022language}. Pretrained on massive text corpora, LLMs learn to generate reliable probability distributions over a predefined vocabulary. This capability allows for the direct application of entropy-based methods to address critical challenges such as hallucinations, where a model generates factually incorrect or nonsensical text~\cite{azaria2023internal,kuhn2023semantic,chen2024context}. Recent studies have empirically demonstrated that incorrectly generated tokens often exhibit higher entropy than correct ones, establishing entropy as a practical proxy for prediction reliability~\cite{xiao2021hallucination, kadavath2022language,chen2024context}.

Multimodal large language models (MLLMs)~\cite{li2023blip,liu2023visual,bai2023qwen,llavaov,chen2024internvl} extend the capabilities of LLMs by integrating visual encoders which process images into visual tokens that are subsequently fed into the model alongside text tokens. While this fusion of modalities enables remarkable new capabilities, MLLMs inherit many of the same uncertainty characteristics as their unimodal predecessors. Consequently, they exhibit similar entropy patterns, where higher entropy in the output distribution often correlates with incorrect or hallucinatory predictions~\cite{leng2024mitigating,zou2024look,wu2025generate, wang2024valid}. This parallel suggests that uncertainty estimation via entropy can be similarly leveraged to enhance the interpretability of MLLM outputs.

While prior work has primarily used entropy for post-hoc hallucination detection, its potential as a proactive signal to guide a model's perceptual process, particularly for complex visual tasks, remains largely underexplored. Therefore, we propose a novel, training-free framework that leverages entropy to enhance MLLM performance on complex visual tasks. Our approach uses entropy not merely as an error detector but as an active guide for decision-making. 

\section{The Inverse Correlation Between Performance and Entropy}\label{supp_sec:correlation}

\begin{figure}[h]
\vspace{-2em}
\centering
\includegraphics[width=1.0\linewidth]{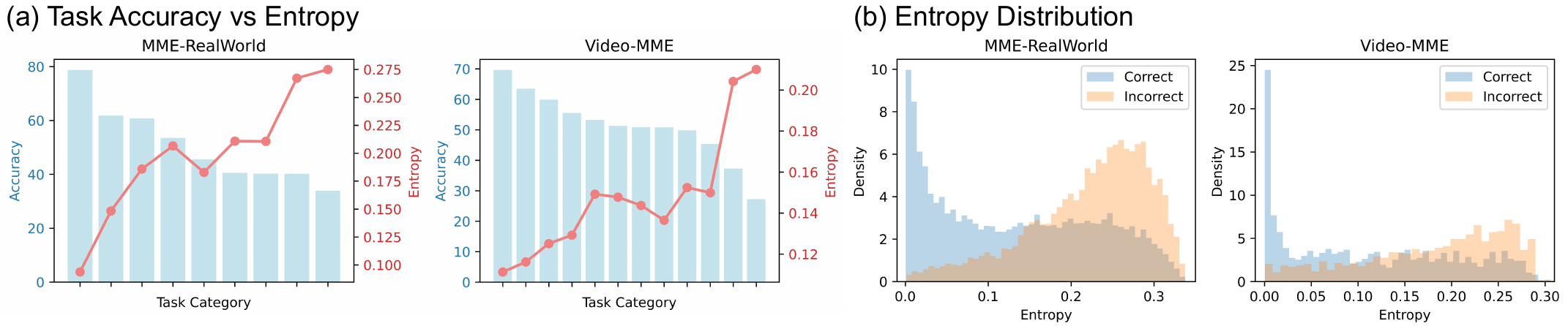}
\caption{(a) Correlation between sub-task accuracy and entropy. (b) Entropy distribution for correct and incorrect predictions.}\label{fig:intro_supp}
\vspace{-2em}
\end{figure}

In this section, we systematically analyze the relationship between MLLM performance and output entropy on demanding multimodal tasks. While prior work has established that incorrectly generated tokens in language models often exhibit higher entropy, this phenomenon has not been thoroughly examined in the context of modern multimodal tasks that require fine-grained understanding of sparse information within noisy visual contexts. We begin by evaluating LLaVA-OneVision 7B~\cite{llavaov} on two challenging benchmarks: Video-MME~\cite{fu2025video} and MME-RealWorld~\cite{zhang2024mme}. MME-RealWorld assesses MLLMs on real-world scenarios across nine categories using 2k-resolution images, while Video-MME features a diverse set of 12 tasks on long-form videos averaging 17 minutes. 

As shown in \cref{fig:intro_supp}a, we plot the accuracy for each sub-task in descending order alongside its corresponding average entropy. In both benchmarks, a clear inverse correlation emerges: sub-tasks with higher accuracy exhibit lower entropy, whereas more difficult sub-tasks yield higher entropy. We also calculate the Pearson Correlation Coefficient~\cite{kirch2008pearson} between entropy and accuracy, achieving -0.90 and -0.95 for MME-RealWorld and Video-MME respectively, which represents strong negative correlation. This finding aligns with principles from curriculum learning, where task difficulty can be measured by model uncertainty~\cite{bengio2009curriculum,kumar2010self}.

Furthermore, \cref{fig:intro_supp}b depicts the entropy distribution for correct versus incorrect predictions. The entropy values for correct predictions are concentrated in a much lower range than those for incorrect ones. This observation confirms that findings from prior work on unimodal hallucination~\cite{leng2024mitigating,zou2024look,wu2025generate, wang2024valid}, where incorrect outputs correlate with higher entropy, also hold for complex multimodal tasks.

\section{Theoretical Discussion of the UG Framework}
\label{supp_sec:theory}

In this section, we provide a formal theoretical grounding for the Uncertainty-Guided (UG) framework and clarify its connection to the Information Bottleneck (IB) principle \cite{tishby2000information, alemi2016deep}. The key finding is that, at inference time, selecting the visual input that \emph{minimizes the model's output uncertainty} serves as a tractable proxy for \emph{maximizing predictive mutual information} between the model's internal representation and the ground-truth output. 

Under this interpretation, the UG framework is viewed as a discrete, inference-time approximation of the IB objective. We show that by varying the input $v$ over a candidate set $\mathcal{V}$, the framework effectively selects the representation $Z = f_\theta(v,q)$ that preserves the most task-relevant information.

\subsection{Notation and Setup}\label{supp_subsec:notation}

Let $V$ and $Q$ denote random variables for visual input and query under the task data distribution, and let $Y_{\text{data}}$ denote the ground-truth output random variable. A fixed multimodal model $f$ with parameters $\theta$ generates a deterministic representation:
\begin{equation}
Z = f_\theta(v, q),
\end{equation}
for an input pair $(v,q)$ where $v \in \mathcal{V}$ and $q \in Q$. Model-generated tokens $Y_{\text{model}}$ are sampled from an autoregressive predictive distribution:
\begin{equation}
p_{i,j}(v,q) = \Pr_\theta\big(Y_{\text{model},i} = \text{token}_j \mid v,q, Y_{\text{model},<i}\big),
\end{equation}
where $p_{i,j}$ denotes the probability at step $i \in \{1,\dots,T\}$ for vocabulary token $j \in \{1,\dots,N\}$.

\subsection{Background: The Information Bottleneck Objective}\label{supp_subsec:ib_background}

The Information Bottleneck principle seeks a representation $Z$ that compresses the input $(v,q)$ while preserving predictive information about the target $Y_{\text{data}}$ \cite{tishby2000information}. For a fixed model $f_\theta$, the IB objective is:
\begin{equation}
\mathcal{L}_{\mathrm{IB}}
=
\underbrace{I\!\big(Z;(v,q)\big)}_{\text{Compression}}
\;-\;
\beta\,\underbrace{I\!\big(Z;Y_{\text{data}}\big)}_{\text{Prediction}}, \label{eq:ib_objective_appendix}
\end{equation}

where $I(\cdot\,;\cdot)$ denotes mutual information and $\beta > 0$ balances compression and prediction terms. High $\beta$ values prioritize prediction (i.e., accuracy), while low values prioritize compression.

In the UG framework, the model parameters $\theta$ are fixed, and we vary only the input $v$ over a finite candidate set $\mathcal{V}$ (e.g., visual crops or frame windows). Since all $v \in \mathcal{V}$ typically share the same dimensionality (i.e., equal size) and are processed by the same fixed-weight encoder, the compression term $I(Z ; (v,q))$, representing the descriptive complexity of the representation, is approximately invariant across candidates. Thus, minimizing the IB objective reduces to maximizing the predictive information:
\begin{equation}
\arg\min_{v \in \mathcal{V}} \mathcal{L}_{\mathrm{IB}} \approx \arg\max_{v \in \mathcal{V}} I(Z ; Y_{\text{data}}).
\end{equation}
Using the decomposition $I(Z ; Y_{\text{data}}) = \mathrm{H}(Y_{\text{data}}) - \mathrm{H}(Y_{\text{data}} \mid Z)$, and noting that $\mathrm{H}(Y_{\text{data}})$ is a constant property of the data distribution, we have:
\begin{equation}
\arg\max_{v \in \mathcal{V}} I(Z ; Y_{\text{data}}) \equiv \arg\min_{v \in \mathcal{V}} \mathrm{H}(Y_{\text{data}} \mid Z).
\label{eq:mi_to_conditional_entropy}
\end{equation}

On the other hand, the UG framework uses the model's predictive (token-level) distributions to compute an \emph{Average Entropy} score:
\begin{equation}
\widehat{\mathcal{H}}(v,q)
\;=\;
-\frac{1}{T}\sum_{i=1}^{T}\sum_{j=1}^{N} p_{i,j}(v,q)\, \log p_{i,j}(v,q),
\label{eq:ug_average_entropy}
\end{equation}
which is proportional to the model's conditional entropy over generated tokens $Y_{\text{model}}$ at $(v,q)$. This quantity \emph{does} depend on $v$ and is tractable at inference time.

\subsection{UG framework as Inference-Time IB Optimization}\label{supp_subsec:main_theorem}

\noindent
\textbf{Calibration and Proxy Assumption.} 
We assume the model's predictive distribution is a calibrated proxy for the conditional data distribution: $\Pr_\theta(Y_{\text{model}} \mid Z) \approx \Pr(Y_{\text{data}} \mid Z)$. Under this assumption, the ordering of candidate inputs by conditional entropy is preserved (i.e., for any $a,b$, $\mathrm{H}(Y_{\text{data}}\!\mid Z_a) < \mathrm{H}(Y_{\text{data}}\!\mid Z_b)$ implies $\widehat{\mathcal{H}}(v_a,q) < \widehat{\mathcal{H}}(v_b,q)$ with high probability). As a result, minimizing the model's output entropy serves as a surrogate for maximizing predictive information. Empirical evidence in \cref{supp_sec:correlation} and main paper analysis suggests that recent MLLMs exhibit reliable calibration unlike conventional neural networks~\cite{guo2017calibration}, which supports this approximation in practice.

\begin{theorem}[UG Framework as Inference-Time Information Bottleneck Optimization]
\label{thm:ug_ib_appendix}
Let $\mathcal{V}$ be a finite set of candidate inputs. At inference time with fixed parameters $\theta$, the IB objective \cref{eq:ib_objective_appendix} is minimized by the candidate that minimizes the conditional entropy of the data given the representation. If the calibration proxy holds such that the ordering of candidates is preserved, then the UG selection rule:
\begin{equation}
v^* = \arg\min_{v_m \in \mathcal{V}} \widehat{\mathcal{H}}(v_m,q)
\end{equation}
is an inference-time approximation to the IB optimizer.
\end{theorem}

\begin{proof}
Since $I(Z; (v,q))$ is constant for fixed-size inputs and fixed $\theta$, the variation in $\mathcal{L}_{\mathrm{IB}}$ is dominated by $- \beta I(Z ; Y_{\text{data}})$. Minimizing this is equivalent to minimizing $\mathrm{H}(Y_{\text{data}} \mid Z)$ per \cref{eq:mi_to_conditional_entropy}. By the calibration assumption, $\widehat{\mathcal{H}}(v_m,q)$ yields the same ranking as $\mathrm{H}(Y_{\text{data}} \mid Z_m)$, completing the approximation.
\end{proof}

\section{Details of Baseline Methods}\label{supp_sec:details_of_baselines} 

In this section, we detail the baseline methods compared to UG methods. We begin with general-purpose MLLMs used in our main experiments, followed by task-specific models (training-free or fine-tuned) compared in our paper.

\subsection{General MLLM Baselines}\label{supp_subsec:general_baselines}
\begin{itemize}
\item\textbf{LLaVA-OneVision}~\cite{llavaov}: An open-source model for unified image and video understanding, connecting a SigLIP encoder with Qwen2 via an MLP projector. Trained on large-scale instruction datasets with AnyRes, it supports scales of 0.5B, 7B, and 72B parameters. 

\item\textbf{Qwen2.5-VL}~\cite{bai2025qwen2}: A vision-language model trained on 4.1T tokens, supporting dynamic resolution and temporal scaling. It excels in object grounding, OCR, and long video understanding. Available in 3B, 7B, 32B, and 72B sizes. 

\item\textbf{InternVL-2.5}~\cite{chen2024expanding}: Built on the InternViT encoder and InternLM2.5/Qwen2.5, this model supports multi-image reasoning and video QA. It employs progressive scaling and dynamic high-resolution strategies (1B to 78B parameters). 

\item\textbf{LLaVA-Video}~\cite{zhang2024video}: An extension of LLaVA-OneVision fine-tuned on LLaVA-Video-178K. It unifies visual representation for images and videos, enabling temporal reasoning (7B and 72B parameters). 

\item\textbf{InternVideo-2.5}~\cite{wang2025internvideo2}: A video-centric MLLM built on InternVL-2.5, enhanced with Long and Rich Context (LRC) modeling. It introduces hierarchical token compression for videos up to 6× longer than previous versions (up to 8B parameters). 
\end{itemize}

\subsection{Visual Search}\label{supp_subsec:baselines_search}
\textbf{Training-free Methods}
\begin{itemize}
\item\textbf{ZoomEye}~\cite{shen2024zoomeye}: A training-free method that represents images as a tree hierarchy. It uses tree search to dynamically zoom into relevant regions. While effective for high-resolution images, its applicability to standard VQA or video tasks remains unproven.

\item\textbf{ViCrop}~\cite{zhang2025mllms}: Exploits internal attention/gradient maps to identify and crop relevant regions. While it boosts fine-grained task performance, improvements are marginal on recent MLLMs (e.g., Qwen2.5-VL), and video application is unexplored. 

\item\textbf{TextCoT}~\cite{luan2024textcot}: A Chain-of-Thought framework for text-rich images, using a three-stage process (overview, coarse localization, fine-grained observation). Its application is primarily limited to text recognition tasks. 
\end{itemize}

\noindent
\textbf{Fine-tuned Methods} 
\begin{itemize}
\item\textbf{SEAL}~\cite{wu2024v}: Introduces MLLM-guided visual search to build a visual working memory. Unlike UG methods, SEAL requires fine-tuning the MLLM to optimize the interaction between search and reasoning. 

\item\textbf{Thyme}~\cite{zhang2025thyme}: Enables MLLMs to perform image manipulations via code. It uses a two-stage training strategy (SFT followed by RL). It improves high-resolution perception but requires complex training pipelines. 
\end{itemize}

\subsection{Video Frame Sampling}\label{supp_subsec:baselines_sample} 

\textbf{VLM-based Methods} 
\begin{itemize}
\item\textbf{BOLT}~\cite{liu2025bolt}: A training-free method using query-guided inverse transform sampling. It relies on CLIP or SigLIP for similarity scores, which inherently lack fine-grained understanding. 

\item\textbf{AKS}~\cite{tang2025adaptive}: Adaptive Keyframe Sampling optimizes relevance and coverage within a token budget. It uses CLIP or BLIP for relevance, which limits fine-grained perception capabilities.
\end{itemize}

\noindent
\textbf{Graph-based Method} 
\begin{itemize}
\item\textbf{KFC}~\cite{fang2025threading}: Selects keyframes by optimizing relevance and diversity, then threads them with textual narratives from a lightweight captioner. This avoids fine-tuning but adds architectural complexity. 
\end{itemize}

\noindent
\textbf{MLLM-based Methods} 
\begin{itemize}
\item\textbf{FRAG}~\cite{huang2025frag}: Independently scores frames for relevance and selects the top-$K$ frames. While it requires no fine-tuning, FRAG incurs high inference time not fully addressed in the original paper. 
\end{itemize}

\subsection{Temporal Grounding}\label{supp_subsec:baselines_ground}  

\textbf{Training-free Methods} 
\begin{itemize}
\item\textbf{VTG-GPT}~\cite{xu2024vtg}: A zero-shot method that debiases queries and converts frames to captions, using a proposal pipeline to predict timestamps.

\item\textbf{TFVTG}~\cite{zheng2024training}: Uses LLMs to decompose queries into sub-events and VLMs to score relevance, integrating results based on temporal logic.

\item\textbf{TAG}~\cite{lee2025tag}: Incorporates temporal pooling, coherence clustering, and similarity adjustment to address semantic fragmentation in zero-shot grounding. 
\end{itemize}

\noindent
\textbf{Instruction-tuned Methods} 

\begin{itemize}
\item\textbf{VTimeLLM}~\cite{huang2024vtimellm}: Uses a three-stage training pipeline (alignment, boundary learning, instruction tuning). It requires extensive fine-tuning, hindering transferability. 

\item\textbf{TimeChat}~\cite{ren2023timechat}: Features a timestamp-aware frame encoder and a sliding Q-Former. While accurate for timestamps, its utility for general VQA or long video tasks is unverified. 

\item\textbf{TimeMarker}~\cite{chen2024timemarker}: Interleaves ``Temporal Separator Tokens'' with frame tokens. The introduction of custom tokens prevents easy application to standard MLLMs. 
\end{itemize}

\section{Details of Benchmarks}\label{supp_sec:details_of_benchmarks}

In this section, we present the benchmarks evaluated in our paper. 

\subsection{Visual Search}\label{supp_subsec:bench_search}

\textbf{Visual Search Benchmarks}
\begin{itemize}
\item \textbf{$V^*$ Bench}~\cite{wu2024v}: $V^*$ Bench consists of 191 questions across two tasks: Attribute Recognition (115 samples) and Spatial Relationship Reasoning (76 samples). It utilizes high-resolution, visually crowded images (around 2k), focusing on fine-grained attribute recognition and spatial reasoning. The questions are multiple-choice; attribute tasks query object properties (color, material), while spatial tasks query relative positions. The predicted option is extracted from the generated output via a rule-based function and matched with the ground truth.

\item \textbf{HR-Bench}~\cite{wang2025divide}: HR-Bench includes ultra-high-resolution images and consists of two splits (4k and 8K), containing 200 images each with different resolutions. Each split is divided into two subtasks: Fine-grained Single-instance Perception (FSP) and Fine-grained Cross-instance Perception (FCP). It tests MLLMs on detailed perception and reasoning tasks, such as attribute recognition, OCR, and spatial relationships. Questions follow a multiple-choice format (A–D). The output string is matched with the ground truth by querying OpenAI gpt-3.5-turbo.

\item \textbf{MME-RealWorld}~\cite{zhang2024mme}: MME-RealWorld is a large-scale benchmark with 13,366 high-resolution images (around 2k) and 29,429 QA pairs, covering 43 subtasks across five real-world domains (Monitoring, OCR, Diagram/Table, Autonomous Driving, and Remote Sensing). Questions are multiple-choice (A–E). Predictions are extracted via rule-based patterns and matched with the ground truth.
\end{itemize}

\noindent
\textbf{Standard VQA Benchmarks}
\begin{itemize}
\item \textbf{DocVQA}~\cite{mathew2021docvqa}: DocVQA evaluates document understanding, containing questions over scanned documents, invoices, and forms. It consists of 5350 examples for validation set and 44650 examples for test set where we adopt validation set to evaluate our models. It assesses the ability to extract and reason over textual and structural information. The output format consists of short text answers, often requiring exact string extraction.

\item \textbf{POPE}~\cite{li2023evaluating}: POPE focuses on hallucination detection. It includes around 9,000 image-question pairs designed to test whether the model hallucinates objects not present in the image. Questions are binary (Yes/No), measuring factual grounding.

\item \textbf{TextVQA}~\cite{singh2019towards}: TextVQA evaluates Optical Character Recognition (OCR) and reasoning over text in natural images. Among train, validation, and test splits, we evaluate on validation set which consists of 5,000 questions. The benchmark tests reading comprehension in visual contexts (signs, labels). Outputs are short text answers, typically spanning one or two words.

\item \textbf{GQA}~\cite{hudson2019gqa}:
GQA measures compositional visual reasoning over real-world images. It contains questions grounded in images from Visual Genome, generated using functional scene-graph programs to ensure high semantic fidelity. The benchmark stresses multi-step reasoning (spatial relations, attributes, comparisons) with reduced dataset biases. Answers are short text phrases, often single words tied to objects or properties in the scene.
\end{itemize}

\subsection{Video Frame Sampling}\label{supp_subsec:bench_sample}

\textbf{Long Video Benchmarks}
\begin{itemize}
\item\textbf{Video-MME}~\cite{fu2025video}: Video-MME evaluates MLLMs on comprehensive video understanding. It consists of 900 videos (254 hours total) with 2,700 annotated QA pairs. Videos span six domains (e.g., knowledge, film, sports) and vary in duration from short (11s–2min) to long (30–60min). It integrates frames, subtitles, and audio to assess cross-modal understanding. Questions are multiple-choice, and predictions are extracted via rule-based parsing.

\item\textbf{MLVU}~\cite{zhou2025mlvu}: MLVU targets videos ranging from 3 minutes to 2 hours. It consists of 3,102 questions (2,593 dev, 509 test) across diverse genres. Tasks include holistic understanding (summarization), single-detail reasoning (plot QA), and multi-detail reasoning (anomaly detection). It tests long-context reasoning and multi-task adaptability using a multiple-choice format.

\item\textbf{LongVideoBench}~\cite{wu2024longvideobench}: LongVideoBench focuses on long-context interleaved video-language understanding. It contains 3,763 videos with subtitles and 6,678 questions across 17 categories. Videos are up to one hour long, and questions follow a “referring reasoning” paradigm requiring retrieval of specific details.
\end{itemize}

\noindent
\textbf{Short Video Benchmarks}
\begin{itemize}
\item\textbf{EgoSchema}~\cite{mangalam2023egoschema}: EgoSchema evaluates egocentric video understanding and commonsense reasoning. It contains $\sim$5,000 short clips paired with multiple-choice questions. The benchmark assesses temporal reasoning and schema-based understanding, often requiring prediction of future actions.

\item\textbf{ActivityNet-QA}~\cite{yu2019activitynet}: ActivityNet-QA is a large-scale video question-answering benchmark built on 5,800 untrimmed ActivityNet videos paired with 58,000 human-annotated QA pairs. It evaluates spatio-temporal understanding and long-term video reasoning, requiring models to interpret complex activities over extended durations and answer free-form or yes/no questions about the depicted events.

\item\textbf{NextQA}~\cite{xiao2021next}: NextQA focuses on next-event prediction and temporal reasoning in short videos. It includes $\sim$51,000 questions over 5,400 videos. Questions (multiple-choice and open-ended) test causal and temporal relationship inference.
\end{itemize}

\subsection{Temporal Grounding}\label{supp_subsec:bench_ground}
\begin{itemize}
\item\textbf{Charades-STA}~\cite{gao2017tall}: Charades-STA benchmarks temporal activity localization. It contains 12,408 clips with 16,128 sentence annotations. The model must predict start and end timestamps for a described activity. Performance is measured by computing the IoU score between the predicted timeline and the ground truth.

\item\textbf{ActivityNet Captions}~\cite{krishna2017dense}: ActivityNet Captions focuses on dense video captioning and grounding. It includes 10,000 videos with $\sim$37,400 queries. The benchmark tests the alignment of multiple natural language descriptions with corresponding temporal segments in long videos.
\end{itemize}

\section{Implementation Details}\label{supp_sec:details}
This section provides a comprehensive overview of the implementation details for our Uncertainty-Guided (UG) framework. To ensure fair and reproducible comparisons, all benchmarks were evaluated using \emph{LMMs-Eval}~\cite{zhang2024lmmseval}. We adopted the library's default evaluation protocol and inference parameters for each benchmark, including temperature, maximum token count, and beam search settings. For further details on the evaluation setup, please refer to the official repository: \url{https://github.com/EvolvingLMMs-Lab/lmms-eval}. The same decoding configuration was used for both the uncertainty scoring phase (Average Entropy or BRC score) and the final answer generation.

A notable implementation detail concerns batch processing. While our uncertainty scoring method is inherently parallelizable as the score for each candidate visual input can be computed independently, we observed a degradation in MLLM performance when using a batch size greater than one. This is a known issue with several open-source MLLMs, where batch inference can lead to less reliable outputs. To ensure maximum accuracy and reproducibility, we therefore set the batch size to one for all experiments, including both scoring and final inference. We anticipate that the runtime of our framework could be substantially reduced with hardware support for true parallel processing or as MLLMs evolve to better support reliable batch inference.

\subsection{UG-Search}\label{supp_subsec:ug_search_impl}

For UG-Search, we employ a sliding window to generate candidate crops from the high-resolution input image. Considering image resolution, we determined the crop size to be one-sixth of the image’s smaller dimension for Visual Search benchmarks and one-half for Standard benchmarks. For each candidate, both the original full-resolution image and the resized crop are provided as input to the MLLM. In \cref{supp_sec:extra_ablation}, we empirically found that resizing the smaller crop back to the original image resolution allows the MLLM to perceive its details more effectively, leading to better performance. During the scoring phase, each input prompt begins with the system prompt and the original image. To avoid redundant computation and improve efficiency, we cache the key–value (KV) representations of system text tokens and original image tokens which are reused for all scoring passes.

The benchmarks used in this task differ in their answer formats. $V^*$ Bench~\cite{wu2024v} and HR-Bench~\cite{wang2025divide} contain multiple-choice questions, whereas TextVQA~\cite{singh2019towards}, POPE~\cite{li2023evaluating}, DocVQA~\cite{mathew2021docvqa}, and GQA~\cite{hudson2019gqa} require short-phrase or yes/no responses. In these settings, the model typically produces a single answer token (e.g., ``A''), followed by an end-of-sequence token ($\langle EOS \rangle$). We compute the average entropy over all generated tokens, including both the answer token and the $\langle EOS \rangle$ token.

\subsection{UG-Sample}\label{supp_subsec:ug_sample_impl}

In UG-Sample, each video frame is treated as an independent candidate visual input, and we use the average entropy of the model’s output to estimate its relevance to the query. Our analysis indicates that using a single-frame window strikes an effective balance between accuracy and computational efficiency. This is because long-video QA tasks typically hinge on identifying sparse, highly informative moments rather than reasoning over extended continuous actions. Following standard practice and considering computational cost~\cite{liu2025bolt,tang2025adaptive}, we select the top 8 frames from a pool of 256 uniformly sampled candidate frames.

We evaluate our method on both multiple-choice and free-form (short-phrase or yes/no) video QA benchmarks, covering long-video datasets such as Video-MME~\cite{fu2025video}, MLVU~\cite{zhou2025mlvu}, and LongVideoBench~\cite{wu2024longvideobench}, as well as short-video datasets including EgoSchema~\cite{mangalam2023egoschema}, ActivityNet-QA~\cite{yu2019activitynet}, and NeXT-QA~\cite{xiao2021next}. Consistent with UG-Search, we compute the average entropy across all generated tokens for scoring.

\subsection{UG-Ground}\label{supp_subsec:ug_ground_impl}
\vspace{-1em}
\begin{algorithm}
\caption{Modified Kadane's Algorithm for Maximum Subarray}\label{alg:kadane}
\begin{algorithmic}[1]
\State \textbf{Input:} A list of BRC scores, $scores$.
\State \textbf{Output:} Indices \texttt{start}, \texttt{end} of maximum sum subarray

\State $max\_sum \gets -\infty$
\State $current\_sum \gets 0$
\State $start \gets 0$, $end \gets 0$, $temp\_start \gets 0$

\For{$i = 0$ \textbf{to} $length(scores) - 1$}
    \If{$current\_sum \leq 0$}
        \State $current\_sum \gets scores[i]$
        \State $temp\_start \gets i$
    \Else
        \State $current\_sum \gets current\_sum + scores[i]$
    \EndIf

    \If{$current\_sum > max\_sum$}
        \State $max\_sum \gets current\_sum$
        \State $start \gets temp\_start$
        \State $end \gets i$
    \EndIf
\EndFor
\State \Return $start$, $end$
\end{algorithmic}
\end{algorithm}
\vspace{-1em}
For UG-Ground, we calculate a Binary Response Confidence (BRC) score for each temporal segment using a sliding window. We prompt the MLLM with a standardized yes/no question for each window:

\begin{quote}
\texttt{Given the action: $\langle$target action$\rangle$, is this action depicted in the video? \\
A. yes \\
B. no \\
Answer with the option's letter from the given choices directly.}
\end{quote}

Here, \texttt{$\langle$target action$\rangle$} is the query event. The BRC score is then calculated from the model's output probabilities for the tokens corresponding to ``A'' and ``B''. This process transforms the video into a sequence of confidence scores, reducing the temporal grounding task to a Maximum Subarray Sum problem. We solve this efficiently in linear time using a modified version of \textit{Kadane’s Algorithm}~\cite{bentley1984programming}, detailed in \cref{alg:kadane}. The algorithm's output (the start and end indices of the maximum sum subarray) is mapped back to timestamps in seconds to produce the final result.

Based on our analysis, which highlighted the importance of perceiving continuous motion for action localization, we use a sliding window of 15 frames with a stride of 1. To manage computational load, videos from Charades-STA~\cite{gao2017tall} were sampled at 3 FPS, while videos from ActivityNet Captions~\cite{krishna2017dense} were sampled at 1 FPS. For baseline comparisons, we restricted the input for the base MLLMs to 64 frames considering their context length limitations.

\section{Extra Ablation Studies}\label{supp_sec:extra_ablation}

This section provides additional ablations to examine how key design choices influence the performance and efficiency of our UG framework.

\subsection{Ablation on Top-$K$ Crops of UG-Search}\label{supp_subsec:topk_ablation}

\begin{table}[h]
  \centering
  \setlength\tabcolsep{3pt}
  \vspace{-1em}
  \caption{\textbf{UG-Search Ablation of Top-$K$ Crops} evaluated on $V^*$ Bench. We ablate Top-$K$ ($K \in \{1,2,3,4,5\}$) for selected visual crops during answering stage. Resizing the crop (Resize) back to the original image resolution generally improves performance.}
  \label{tab:supp_search_ablation}
  \begin{adjustbox}{width=0.8\textwidth}
  \begin{tabular}{lccc}
    \toprule
    Method & InternVL2.5-8B & Qwen2.5-VL-7B & LLaVA-OV-7B \\
    \midrule
    Baseline & 71.7 & 65.5 & 74.4 \\
    \ours{\quad w/ Top-1 w/o Resize} & \ours{83.3} & \ours{77.5} & \ours{82.2} \\
    \ours{\quad w/ Top-1} & \ours{83.3} & \ours{79.1} & \ours{\textbf{83.3}} \\
    \ours{\quad w/ Top-2} & \ours{83.3} & \ours{80.6} & \ours{80.6} \\
    \ours{\quad w/ Top-3} & \ours{83.8} & \ours{80.6} & \ours{77.0} \\
    \ours{\quad w/ Top-4} & \ours{\textbf{84.8}} & \ours{78.0} & \ours{10.0} \\
    \ours{\quad w/ Top-5} & \ours{82.7} & \ours{\textbf{81.7}} & \ours{10.0} \\
    \bottomrule
  \end{tabular}
  \end{adjustbox}
    \vspace{-1em}
\end{table}

We study two key factors in UG-Search: the number of low-entropy crops used during the answering stage and whether these crops are resized to the model’s original input resolution. As shown in \cref{tab:supp_search_ablation}, resizing provides a clear benefit: for instance, LLaVA-OV-7B improves from 82.2\% (Top-$1$ w/o resizing) to 83.3\% when the same crop is upsampled, suggesting that consistent visual scale helps MLLMs reason over fine-grained details.

Increasing $K$ from one to multiple crops further improves accuracy on $V^*$ Bench. InternVL2.5-8B attains its best score of 84.8\% with Top-$4$ crops, while Qwen2.5-VL-7B peaks at 81.7\% with Top-$5$ crops. These gains indicate that aggregating several high-confidence regions enhances coverage of both localized attributes and relational cues across spatially separated objects. However, some models struggle with many simultaneous inputs (e.g., LLaVA-OV-7B degrades beyond Top-$3$), likely due to limited training on multi-image contexts.

Although multi-crop selection alleviates cases where no single crop contains all relevant objects (see \cref{fig:supp_qualitative_ug_search}), the Top-$1$–with-resizing configuration offers the most reliable balance of accuracy, robustness, and efficiency across all models. We therefore adopt Top-$1$ with resizing as the default setting in our main experiments.

\subsection{Ablation on Scoring Methods}
\label{supp_subsec:scoring_ablation}

\vspace{-1em}

\begin{table}[h]
\centering
\caption{\textbf{Ablation on scoring methods} using LLaVA-OV-7B. The \textbf{bold} number denotes the best result.}
\label{tab:scoring_comparison}
\setlength\tabcolsep{5pt}
\resizebox{0.8\linewidth}{!}{%
\begin{tabular}{lccccc}
\toprule
 & Baseline & Entropy (UG) & Binary & Confidence & SigLIP \\
\midrule
$V^*$ Bench & 74.4 & \textbf{86.9} & 85.9 & 71.7 & 79.6 \\
Video-MME    & 53.9 & \textbf{58.6} & 57.8 & 52.2 & 56.1 \\
\bottomrule
\end{tabular}%
}
\end{table}

\vspace{-2em}

\begin{table}[h]
\centering
\caption{\textbf{Prompt templates for each scoring method.} SigLIP computes similarity directly and requires no prompt.}
\label{tab:scoring_prompt}
\resizebox{\linewidth}{!}{%
\begin{tabular}{p{0.22\linewidth} p{0.39\linewidth} p{0.39\linewidth}}
\toprule
\textbf{Entropy} & \textbf{Binary} & \textbf{Confidence} \\
\midrule
{\ttfamily\small
  \textless image\textgreater{}
  \textless Crop\textgreater{}
  \textless Query\textgreater}
&
{\ttfamily\small
  \textless Crop/Frame\textgreater{}
  Given the question: \textless Query\textgreater,
  is this question relevant to
  the image? A.~yes\enspace B.~no.}
&
{\ttfamily\small
  \textless Crop/Frame\textgreater{}
  \textless Query\textgreater{}
  Answer the question based on
  the image. Then, give your
  confidence from 0 (low) to 10 (high).}
\\
\bottomrule
\end{tabular}%
}
\end{table}

\vspace{-1em}

We compare four uncertainty scoring variants against LLaVA-OV-7B baseline: our entropy-based score (UG), a binary relevance prompt, a confidence self-rating prompt, and an external SigLIP similarity score. As shown in \cref{tab:scoring_comparison}, entropy outperforms all alternatives on both $V^*$ Bench and Video-MME. The confidence-based score notably degrades below the baseline on $V^*$ Bench, consistent with the well-documented overconfidence bias of LLMs~\cite{xiong2024can,tian2023just}: high self-reported confidence scores are unreliable proxies for candidate relevance. The prompt templates used for the three prompt-based variants are provided in \cref{tab:scoring_prompt}; SigLIP requires no prompt as it computes frame–query similarity directly.

\subsection{Precision of the Entropy Signal}
\label{supp_subsec:entropy_precision}
\vspace{-1em}
\begin{table}[h]
\centering
\caption{\textbf{Top-1 precision (\%) of each scoring method.} Precision measures how often the highest-scoring crop or frame contains the ground-truth target. \textbf{Bold} denotes the best result.}
\setlength\tabcolsep{5pt}
\label{tab:entropy_precision}
\resizebox{0.7\linewidth}{!}{%
\begin{tabular}{lcccc}
\toprule
Dataset & Entropy (UG) & Binary & Confidence & SigLIP \\
\midrule
$V^*$ Bench             & \textbf{79.6} & 77.1 & 0.0 & 65.9 \\
$\text{Video-MME}_{kfs}$ & \textbf{44.7} & 41.3 & 4.2 & 36.6 \\
\bottomrule
\end{tabular}%
}
\end{table}
\vspace{-1em}
To directly validate that entropy selects visually informative candidates, we measure the top-1 precision of each scoring method: the fraction of queries for which the highest-scoring crop or frame contains the ground-truth target region. We use key frame annotations from KFS-Bench~\cite{li2026kfs} to construct ground-truth labels for a subset of V-MME ($\text{V-MME}_{kfs}$). As shown in \cref{tab:entropy_precision}, entropy achieves the highest precision on both datasets, confirming that minimizing output entropy is an effective localization signal. Confidence scoring achieves near-zero precision (0.0\% on $V^*$ Bench, 4.2\% on $\text{V-MME}_{kfs}$), directly corroborating the overconfidence analysis in \cref{supp_subsec:scoring_ablation}: a model that assigns uniformly high confidence to all candidates provides no discriminative signal for selection.

\subsection{Comparison with Test-Time Compute Baselines}
\label{supp_subsec:test_time_compute}

\vspace{-1em}

\begin{figure}[h]
\centering
\includegraphics[width=\linewidth]{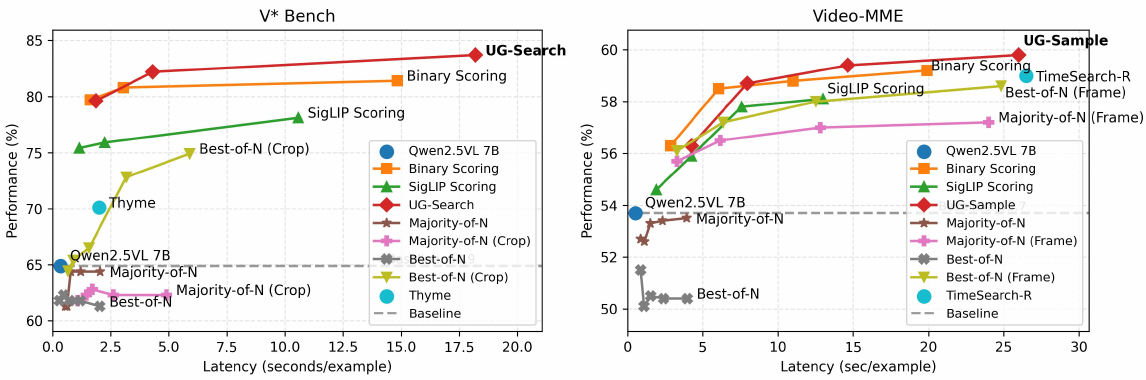}
\caption{\textbf{Performance--latency trade-off against test-time compute and tool-use baselines} on $V^*$ Bench and Video-MME. UG variants are shown alongside Majority-of-N, Best-of-N, and their input-randomized counterparts, as well as RL-trained tool-use models Thyme~\cite{zhang2025thyme} and TimeSearch-R~\cite{pan2025timesearch}.}
\label{fig:test_time_comp}
\end{figure}
\vspace{-1em}

\cref{fig:test_time_comp} compares the performance--latency trade-off of UG against test-time compute baselines and tool-use models. We evaluate two output-side baselines on a fixed input: \emph{Majority-of-N}, which samples $N$ answers and selects by majority vote, and \emph{Best-of-N}, which selects the answer with the lowest output entropy. We also evaluate input-randomized variants, \emph{Majority-of-N (Crop/Frame)} and \emph{Best-of-N (Crop/Frame)}, which sample $N$ random crops or frames and apply the same selection rules, matching the number of forward passes used by UG. All four variants sweep $N \in \{4, 8, 16, 32, 64\}$. For UG, we sweep either the crop size $\mathit{CropSize} \in \{1/2, 1/3, 1/6\}$ with Top-1 selection, or the candidate frame pool $\mathit{CandidateFrames} \in \{32, 64, 128, 256\}$ with Top-8 selection. We additionally compare against RL-trained tool-use models, Thyme~\cite{zhang2025thyme} and TimeSearch-R~\cite{pan2025timesearch}, which invoke a fixed number of tool calls and cannot adaptively adjust their compute budget. Despite requiring no task-specific training, UG achieves a superior performance--latency trade-off over all baselines across both benchmarks.

\section{Extra Results of UG Framework}\label{supp_sec:extra_results}

This section provides additional experimental results that further demonstrate the versatility and effectiveness of our UG framework.

\subsection{Extra Results with UG-Search}\label{supp_subsec:extra_ug_search}
\vspace{-1em}
\begin{table}[h]
  \centering
  \setlength\tabcolsep{3pt}
  \renewcommand{\arraystretch}{1.1}
  \scriptsize
  \caption{\textbf{UG-Search on the MME-RealWorld Benchmark.} Our training-free method significantly boosts perception performance, achieving results competitive with the fine-tuned Thyme model.}
  \label{tab:supp_mmerealworld}
  \resizebox{0.5\columnwidth}{!}{%
    \begin{tabular}{lccc}
      \toprule
      \multirow{2}{*}{\textbf{Model}} & \multicolumn{3}{c}{\textbf{MME-RealWorld}} \\
      & Reason. & Percep. & Overall \\
      \midrule
      QwenVL2.5-7B                 & 33.4 & 51.5 & 49.3 \\
      \quad w/ TextCoT~\cite{luan2024textcot}             & 33.7 & 53.3 & 50.9 \\
      \quad w/ ViCrop~\cite{zhang2025mllms}            & 34.2 & 53.4 & 51.1 \\
      \quad w/ Thyme~\cite{zhang2025thyme}              & \textbf{43.7} & \textbf{60.1} & \textbf{58.1} \\
      \ours{\quad w/ UG-Search}    & \ours{35.1} & \ours{58.5} & \ours{55.7} \\
      \bottomrule
    \end{tabular}
  }
\end{table}
\vspace{-1em}
\noindent
\textbf{UG-Search on MME-RealWorld.}
MME-RealWorld~\cite{zhang2024mme} is a recently introduced benchmark that evaluates multimodal models in practical scenarios, with a particular focus on perception and reasoning skills. As shown in \cref{tab:supp_mmerealworld}, our training-free method provides a substantial boost to the model's perception score (+7.0\%), bringing its overall performance to 55.7 (+6.4\%). This result significantly surpasses other training-free methods like TextCoT~\cite{luan2024textcot} and Rel-att~\cite{zhang2025mllms} and is highly competitive with the fine-tuned Thyme model~\cite{zhang2025thyme}. These findings underscore that our uncertainty-guided approach is a powerful and efficient method for enhancing MLLM performance in practical, real-world scenarios.
\vspace{-1em}
\begin{table}[h]
  \centering
  \setlength\tabcolsep{3pt}
  \renewcommand{\arraystretch}{1.1}
  \scriptsize
  \caption{\textbf{UG-Search with Visual Prompt on $V^*$ Bench.} We tested various Crop Size $\in\{1/2, 1/3, 1/6\}$ with and without visual prompt.}
  \label{tab:supp_vprompt}
  \resizebox{0.7\columnwidth}{!}{%
    \begin{tabular}{lccc}
      \toprule
      \multirow{2}{*}{\textbf{Method}} & \multicolumn{3}{c}{\textbf{$V^*$ Bench}} \\
      & Attribute & Spatial & Overall \\
      \midrule
      LLaVA-OV 7B                       & 79.1 & 67.1 & 74.4 \\
      \ours{\quad w/ Crop Size=$1/2$}   & \ours{85.2} & \ours{75.0} & \ours{81.2} \\
      \ours{\quad w/ Crop Size=$1/2$ VPrompt} & \ours{84.4} & \ours{73.7} & \ours{80.1} \\
      \ours{\quad w/ Crop Size=$1/3$}   & \ours{92.2} & \ours{69.7} & \ours{83.3} \\
      \ours{\quad w/ Crop Size=$1/3$ VPrompt} & \ours{92.2} & \ours{69.7} & \ours{83.3} \\
      \ours{\quad w/ Crop Size=$1/6$}   & \ours{94.8} & \ours{\textbf{75.0}} & \ours{\textbf{86.9}} \\
      \ours{\quad w/ Crop Size=$1/6$ VPrompt} & \ours{\textbf{95.7}} & \ours{73.7} & \ours{\textbf{86.9}} \\
      \bottomrule
    \end{tabular}
  }
\end{table}

\noindent
\textbf{UG-Search with Visual Prompt.}
During the entropy-based scoring stage of UG-Search, we use the standard pre-prompts and post-prompts provided by each benchmark (e.g., ``Answer with the option's letter from the given choices directly.'') together with the task query (e.g., ``What is the color of the truck?''). Inspired by prior work on visual prompting~\cite{shtedritski2023does,shen2024zoomeye}, we additionally evaluate a variant where a red rectangle is drawn on the original image to explicitly indicate the crop region. The prompt is then updated to describe the relationship between the main image and the cropped region. Specifically, we compare:

\begin{itemize}
    \item \textbf{UG-Search template:} \\
    \texttt{<original image> <visual crop> <Query>}
    \item \textbf{UG-Search with VPrompt template:} \\
    \texttt{<original image with red rectangle> <visual crop> First image \\ is the main image, and the red rectangle marks the focused \\ region shown in the second image. <Query>}
\end{itemize}

\cref{tab:supp_vprompt} reports the performance of LLaVA-OneVision-7B on the $V^*$ Bench across crop sizes $\{1/2, 1/3, 1/6\}$. For crop sizes $1/2$ and $1/3$, the VPrompt and original templates yield nearly identical results. For crop size $1/6$, VPrompt slightly improves attribute accuracy (95.7 vs.\ 94.8), while spatial accuracy remains comparable (73.7 vs.\ 75.0). Across all settings, the performance gap between the standard template and VPrompt is minimal, indicating that UG-Search 
is robust to prompt format across varying visual granularities.

\subsection{Extra Results with UG-Sample}\label{supp_subsec:extra_ug_sample}
\vspace{-1em}
\begin{table}[h]
\centering
\setlength\tabcolsep{3pt}
\caption{\textbf{UG-Sample on Video-MME with Subtitles} The results demonstrate the robustness of our method on Video-MME with subtitles. The value in subscript ($\Delta$) indicates the performance change from the baseline (green for gains, red for losses).}
\label{tab:videomme_w_subtitle}
\begin{adjustbox}{width=0.6\textwidth}
{
\begin{tabular}{l|cccc}
\toprule
\multirow{2}{*}{\textbf{Model}}  & \multicolumn{4}{c}{\textbf{Video-MME} \tiny{(w/ sub.)}} \\
                                & Overall     & Short        & Medium      & Long    \\ 
\textit{Video Length}  & \textit{17min} & \textit{1.3min} & \textit{9min}  & \textit{41min} \\
\midrule
LLaVA-OV-7B    & 58.8 & 70.8  & 56.1 & 49.4  \\
\ours{\quad w/ \textbf{UG-Sample}$_\Delta$} & \ours{60.6$_{\textcolor{darkgreen}{1.8}}$} & \ours{72.7$_{\textcolor{darkgreen}{1.9}}$} & \ours{58.8$_{\textcolor{darkgreen}{2.7}}$} & \ours{50.3$_{\textcolor{darkgreen}{0.9}}$} \\
LLaVA-Video-7B    & 67.6 & 71.9  & 64.9 & 66.0  \\
\ours{\quad w/ \textbf{UG-Sample}$_\Delta$} & \ours{69.7$_{\textcolor{darkgreen}{2.1}}$} & \ours{\textbf{74.8}$_{\textcolor{darkgreen}{2.9}}$} & \ours{\textbf{68.3}$_{\textcolor{darkgreen}{3.4}}$} & \ours{66.1$_{\textcolor{darkgreen}{0.1}}$} \\
InternVL2.5-8B   & 61.1 & 70.9  & 60.7 & 51.9  \\
\ours{\quad w/ \textbf{UG-Sample}$_\Delta$} & \ours{\textbf{61.7}$_{\textcolor{darkgreen}{0.6}}$} & \ours{69.2$_{\textcolor{darkred}{1.7}}$} & \ours{62.1$_{\textcolor{darkgreen}{1.4}}$} & \ours{\textbf{53.8}$_{\textcolor{darkgreen}{1.9}}$} \\
InternVideo2.5-8B  & 58.5 & 67.1  & 57.8 & 50.7  \\
\ours{\quad w/ \textbf{UG-Sample}$_\Delta$} & \ours{58.7$_{\textcolor{darkgreen}{0.2}}$} & \ours{68.2$_{\textcolor{darkgreen}{1.1}}$} & \ours{57.0$_{\textcolor{darkred}{0.8}}$} & \ours{50.9$_{\textcolor{darkgreen}{0.2}}$} \\
\bottomrule                       
\end{tabular}
}
\end{adjustbox}
\end{table}
\vspace{-1em}
\noindent
\textbf{Robustness to Linguistic Cues.} 
In \cref{tab:videomme_w_subtitle}, we evaluate the robustness of UG-Sample to the presence of linguistic cues by testing it on Video-MME with subtitles enabled. The results show that UG-Sample continues to deliver consistent performance gains even when subtitles are provided. Although the improvements are smaller compared to the no-subtitle setting, this reduction is expected: subtitles often contain information directly relevant to the query, reducing the relative importance of visual sampling. Nevertheless, the persistent gains indicate that our visually driven, uncertainty-based sampling strategy remains effective and is not rendered redundant by additional textual signals. For example, LLaVA-Video-7B still achieves a +2.1\% overall improvement, including a +3.4\% gain on medium-length videos.

\begin{figure}[h]
\centering
\includegraphics[width=0.9\linewidth]{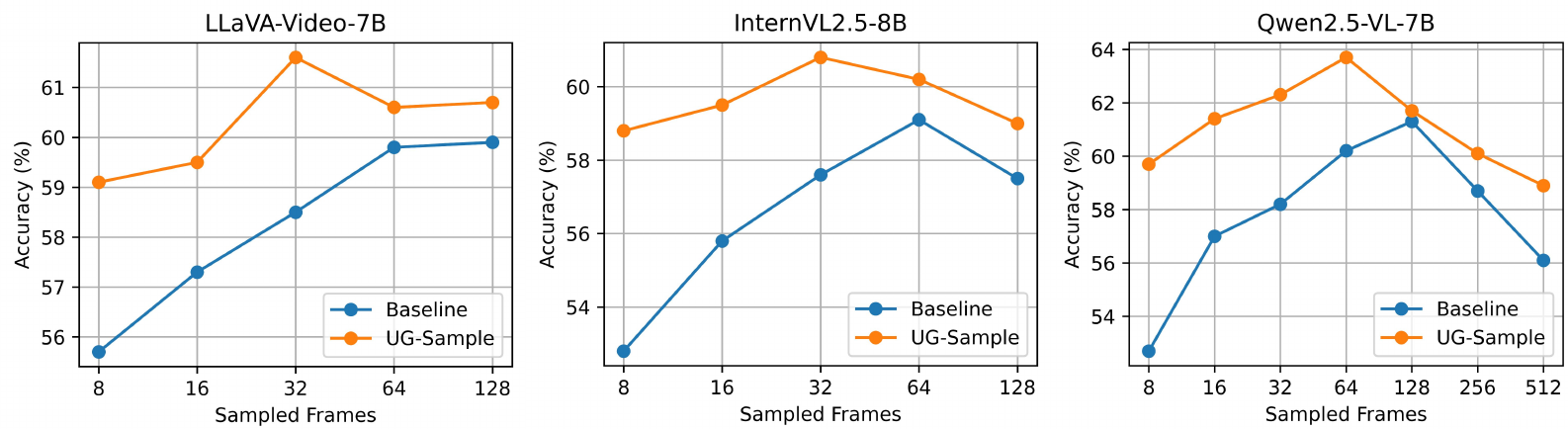}
\caption{\textbf{Scaling UG-Sample to Maximum Frame Budgets} evaluated on LongVideoBench. 
We evaluate LLaVA-Video-7B, InternVL2.5-8B, and Qwen2.5-VL-7B by increasing the number of selected frames from the standard 8-frame setting to the maximum frame count permitted by each model's context length limit.}\label{supp_fig:frame_scale_up}
\end{figure}

\noindent
\textbf{Scaling UG-Sample to Maximum Frame Budgets.}
\cref{supp_fig:frame_scale_up} illustrates the scalability of UG-Sample when increasing the number of selected frames from the standard 8-frame setting used in the main paper to the maximum context length supported by each model. The conventional choice of 8 frames follows prior frame-sampling protocols~\cite{yu2024frame,liu2025bolt,tang2025adaptive}, where 8 frames are sampled from a fixed candidate pool (typically 128 frames or 1 FPS), ensuring a controlled comparison of sampling strategies rather than simply exploiting larger frame counts. To assess whether UG-Sample remains effective in extended regimes, we evaluate both uniform sampling and UG-Sample under progressively larger frame budgets (16, 32, 64, 128, and up to the full context limit). For UG-Sample, candidate frames are obtained at 2~FPS, and scoring is performed using non-overlapping 5-frame windows with a stride of 5 to maintain computational efficiency. For
top-$K$ frame selection during answer generation, we take the center frame of each window (e.g., the 3th frame in a 5-frame window).

The models used in this study have differing tokenization constraints, which determine their maximum feasible frame counts. LLaVA-Video-7B processes 196 visual tokens per frame and reaches its 32k-token limit at roughly 128 frames, after which performance degrades sharply. InternVL2.5-8B uses 64 tokens per frame and fits up to 512 frames within its context, though out-of-memory errors arise beyond 256 frames. Qwen2.5-VL-7B dynamically allocates visual tokens according to the maximum number of pixles ($max\_pixels$), and we adopt its default video configuration (768$\times$28$\times$28) during evaluation.

Across all three models, UG-Sample consistently outperforms uniform sampling for every tested frame count. Notably, UG-Sample with only 16 frames already matches or exceeds the accuracy achieved by uniform sampling even at very large frame budgets, including settings that fully saturate the models' context lengths. As shown in \cref{supp_fig:frame_scale_up}, performance gains are most pronounced in the low-to-mid frame regime (8–64 frames), where UG-Sample leverages uncertainty-driven selection to identify informative segments that uniform sampling cannot reliably capture. These results demonstrate that UG-Sample scales robustly with increasing frame availability, and remains effective even when operating near or at the maximum context capacity of modern multimodal LLMs.

\section{Extra Qualitative Results}\label{supp_sec:qualitative}
\begin{figure}[h] 
\centering
\includegraphics[width=1.0\linewidth]{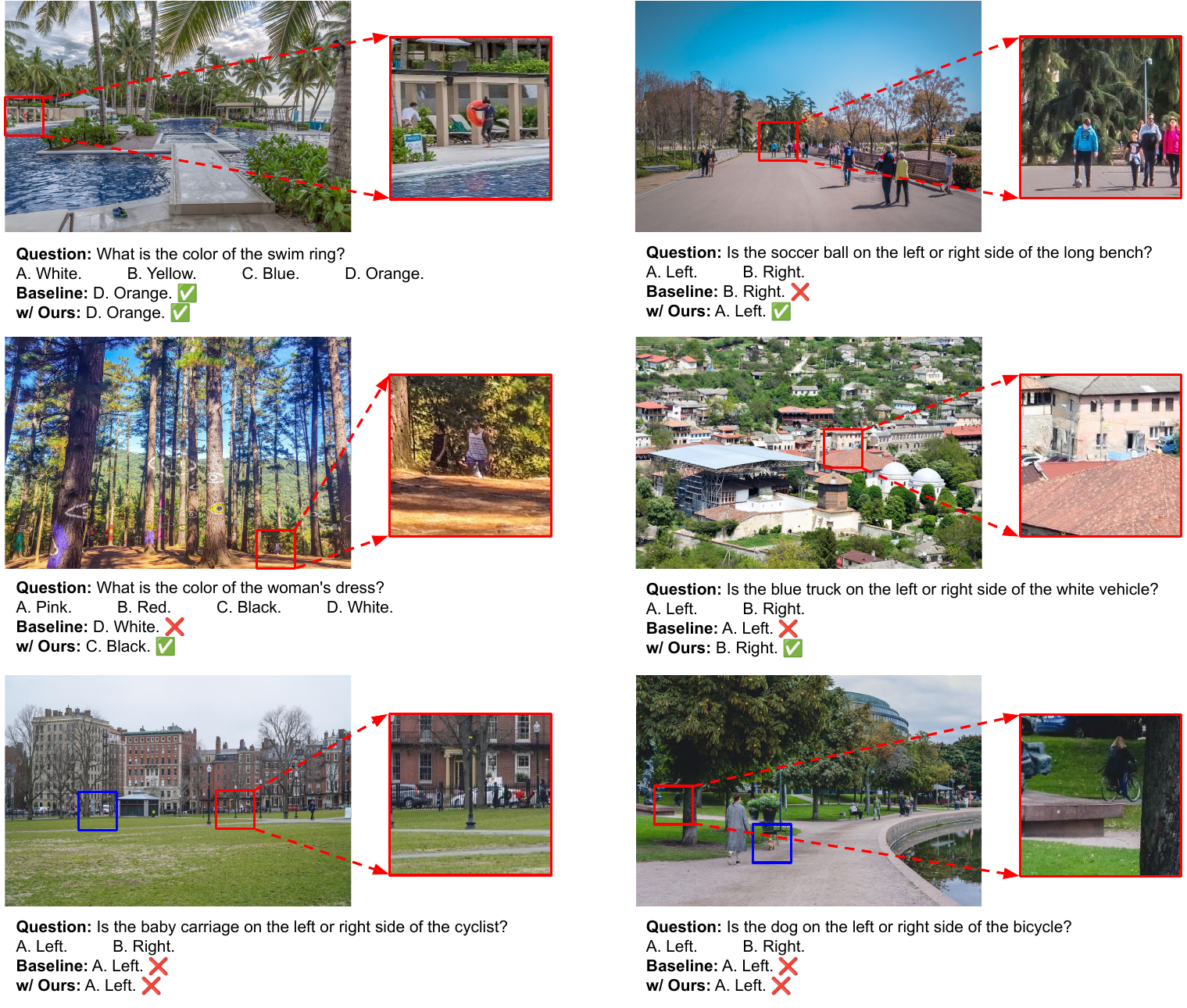}
\caption{\textbf{Qualitative Results of UG-Search.} We compare qualitative results from UG-Search (ours) and InternV2.5-8B (baseline) on $V^*$ Bench. Red rectangle expresses the visual crop with the lowest entropy that is selected by UG-Search. UG-Search successfully capture the relevant object in examples of the first two rows while the last row exemplifies the failure cases where UG-Search capture one object and miss the another. Missed object is marked with blue rectangle.}\label{fig:supp_qualitative_ug_search}
\end{figure}

\begin{figure}[h] 
\centering
\includegraphics[width=1.0\linewidth]{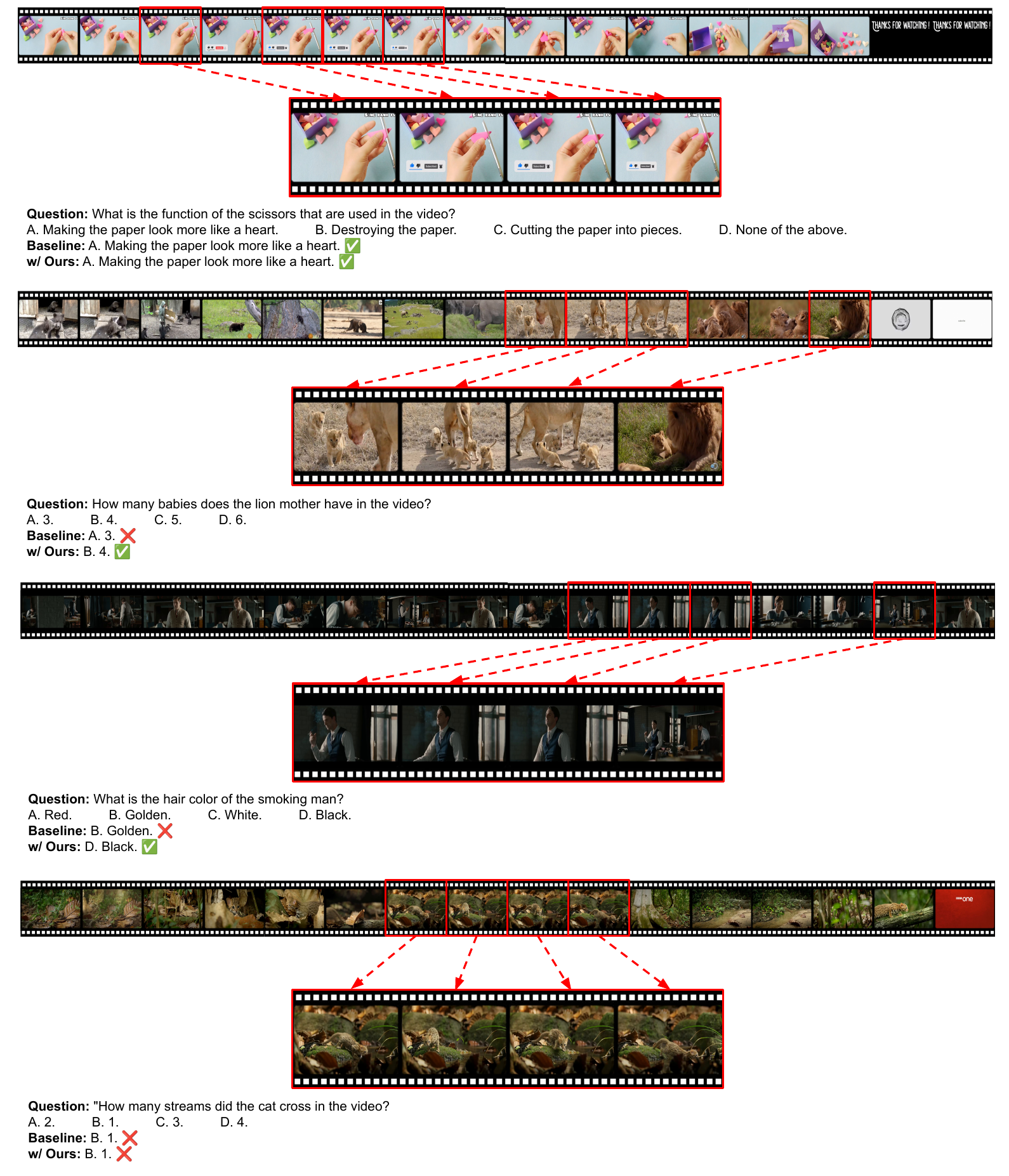}
\caption{\textbf{Qualitative Results of UG-Sample.} We compare qualitative results from UG-Sample (ours) and InternV2.5-8B (baseline) on Video-MME. Red rectangle expresses the frames selected by UG-Sample. We provide a zoom-in version of the selected frames for better visualization. Selected frames are combined into a single context, and used for final inference to answer the query.}\label{fig:supp_qualitative_ug_sample}
\end{figure}

\begin{figure}[h] 
\centering
\includegraphics[width=1.0\linewidth]{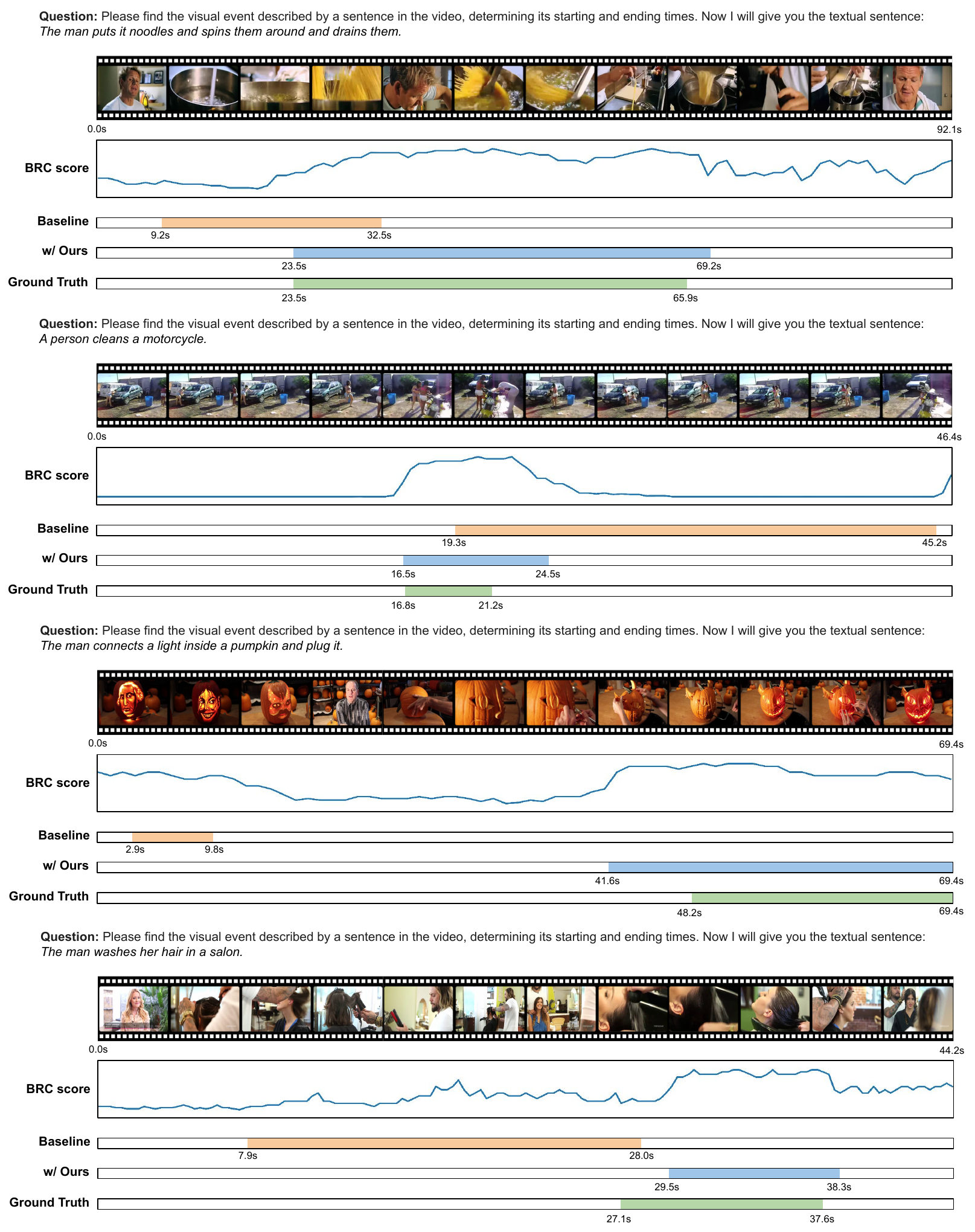}
\caption{\textbf{Qualitative Results of UG-Ground.} We compare qualitative results from UG-Ground (ours) and InternV2.5-8B (baseline) on ActivityNet Captions. The orange, blue and green bars show the grounding results of baseline, UG-Ground, and ground truth respectively. Our UG-Ground method transforms the video into a sequence of BRC scores and then find the subarray with the maximum sum.}\label{fig:supp_qualitative_ug_ground}
\end{figure}

In addition to the examples provided in the main paper, this section presents further qualitative results for each of our three tasks in \cref{fig:supp_qualitative_ug_search}, \cref{fig:supp_qualitative_ug_sample}, and \cref{fig:supp_qualitative_ug_ground}.

\noindent
\textbf{Visual Search.}
\cref{fig:supp_qualitative_ug_search} further demonstrates the ability of UG-Search to overcome the perceptual limitations of the baseline model in both spatial reasoning and attribute recognition. For relational queries, such as determining the relative position of two objects, the baseline model frequently fails. In contrast, our method correctly localizes the relevant objects (e.g., the soccer ball and the bench) enabling it to resolve the spatial relationship accurately. Similarly, for attribute-based questions like identifying the color of a woman's dress, UG-Search successfully focuses on the correct region and provides the right answer where the baseline is misled by other elements in the scene. These examples highlight how uncertainty-guided focus directly translates to more reliable, fine-grained perception. The last row indicates cases where both the baseline and UG-Search fail, specifically when the target objects are small and relatively far apart. In such cases, the fixed bounding box can capture one object but may miss the other, resulting in an incorrect answer.

\noindent
\textbf{Video Frame Sampling.}
The examples in \cref{fig:supp_qualitative_ug_sample} illustrate how UG-Sample mitigates both perceptual and temporal reasoning errors. In a fine-grained visual query about a man's hair color, the baseline model provides an incorrect answer, likely due to a poor selection of frames. Our method, however, correctly identifies the key frames where the attribute is clearly visible (e.g., hair color of the smoking man). For temporal reasoning tasks, such as counting the number of lion cubs appearing over time, UG-Sample selects a more comprehensive set of frames, leading to an accurate count where the baseline underestimates the total. Certain challenging cases remain difficult for both methods, such as counting the streams crossed by a cat, which cannot be solved solely by spatial reasoning. Nevertheless, the overall trend demonstrates that our approach yields more reliable visual and temporal understanding than the baseline.

\noindent
\textbf{Video Temporal Grounding.} \cref{fig:supp_qualitative_ug_ground} provides a clear visualization of UG-Ground's prediction. The baseline model's predictions are often misaligned, either starting too late, ending too early, or drifting entirely off the target event. For example, in the first example asking about ``noodles'', the baseline predicts 9.2s–32.5s, which mostly misses the correct window of 23.5s–65.9s, while our method successfully recovers the full range. Similarly, in the ``cleaning a motorcycle'' example, the baseline's predicted segment is too long and poorly aligned. In contrast, our method's prediction, guided by the distinct peak in the BRC score sequence, aligns almost perfectly with the ground truth. This pattern holds across multiple examples (e.g., ``lighting the pumpkin'' and ``washing hair in a salon''), demonstrating that our uncertainty-based scoring effectively transforms the complex task of temporal grounding into a robust maximum subarray problem, yielding consistently accurate localization.

\clearpage

\end{document}